\documentclass{article}

\usepackage{microtype}
\usepackage{graphicx}
\usepackage{subfigure}
\usepackage{booktabs} 

\usepackage{hyperref}

\usepackage[accepted]{icml2024}

\usepackage{amsmath}
\usepackage{amssymb}
\usepackage{mathtools}
\usepackage{amsthm}

\usepackage{times}
\usepackage{xcolor}
\usepackage{latexsym}
\usepackage{url}
\usepackage{pifont}
\usepackage{subcaption}
\usepackage{multirow}
\usepackage{tcolorbox}

\def\bng{\bngx}

\font\bngx=bang10

\def\*#1*#2{o\null{#2}{#1}}

\def\sh#1{\setbox0=\hbox{#1}%
     \kern-.02em\copy0\kern-\wd0
     \kern.04em\copy0\kern-\wd0
     \kern-.02em\raise.0433em\box0 }
\usepackage{microtype}

\usepackage[capitalize,noabbrev]{cleveref}

\theoremstyle{plain}
\newtheorem{theorem}{Theorem}[section]

\theoremstyle{definition}
\newtheorem{definition}[theorem]{Definition}

\theoremstyle{remark}

\usepackage[textsize=tiny]{todonotes}

\usepackage{amsmath,amsfonts,bm}

\def\vtheta{{\bm{\theta}}}

\def\vc{{\bm{c}}}

\def\vm{{\bm{m}}}

\def\vs{{\bm{s}}}

\def\vv{{\bm{v}}}

\def\vx{{\bm{x}}}

\def\vz{{\bm{z}}}

\DeclareMathAlphabet{\mathsfit}{\encodingdefault}{\sfdefault}{m}{sl}
\SetMathAlphabet{\mathsfit}{bold}{\encodingdefault}{\sfdefault}{bx}{n}

\def\gD{{\mathcal{D}}}

\def\gK{{\mathcal{K}}}

\def\gM{{\mathcal{M}}}

\def\gU{{\mathcal{U}}}
\def\gV{{\mathcal{V}}}

\def\sR{{\mathbb{R}}}

\icmltitlerunning{LESS: Selecting Influential Data for Targeted Instruction Tuning}

\begin{document}

\twocolumn[
\icmltitle{LESS: Selecting Influential Data for Targeted Instruction Tuning}



\icmlsetsymbol{equal}{*}

\begin{icmlauthorlist}
\icmlauthor{Mengzhou Xia}{yyy,equal}
\icmlauthor{Sadhika Malladi}{yyy,equal}
\icmlauthor{Suchin Gururangan}{comp}
\icmlauthor{Sanjeev Arora}{yyy}
\icmlauthor{Danqi Chen}{yyy}
\end{icmlauthorlist}

\icmlaffiliation{yyy}{Princeton Language and Intelligence (PLI), Princeton University, USA}
\icmlaffiliation{comp}{Department of Computer Science, University of Washington, USA}

\icmlcorrespondingauthor{Mengzhou Xia, Sadhika Malladi}{{mengzhou, smalladi}@princeton.edu}
\icmlkeywords{Machine Learning, ICML}

\vskip 0.3in
]



\printAffiliationsAndNotice{\icmlEqualContribution} 

\newcommand{\draftonly}[1]{#1}
\newcommand{\draftcomment}[3]{\draftonly{\textcolor{#2}{{{[#1: #3]}}}}}
\newcommand{\mz}[1]{\draftcomment{Mengzhou}{purple}{#1}}
\newcommand{\sm}[1]{\draftcomment{Sadhika}{blue}{#1}}
\newcommand{\danqi}[1]{\draftcomment{Danqi}{orange}{#1}}
\newcommand{\suchin}[1]{\draftcomment{Suchin}{red}{#1}}

\newcommand{\cmark}{\ding{51}}%
\newcommand{\xmark}{\ding{55}}%
\newcommand{\flan}{\textsc{Flan V2}}
\newcommand{\cott}{\textsc{CoT}}
\newcommand{\dolly}{\textsc{Dolly}}
\newcommand{\bbh}{\textsc{BBH}}
\newcommand{\humaneval}{\textsc{HumanEval}}
\newcommand{\oasst}{\textsc{Open Assistant 1}}
\newcommand{\gsm}{\textsc{GSM}}
\newcommand{\mmlu}{\textsc{MMLU}}
\newcommand{\tydiqa}{\textsc{TydiQA}}
\newcommand{\gptalpaca}{\textsc{GPT4-Alpaca}}
\newcommand{\llamasmall}{\textsc{Llama-2-7B}}
\newcommand{\llamalarge}{\textsc{Llama-2-13B}}
\newcommand{\mistral}{\textsc{Mistral-7B}}
\newcommand{\llamasmallchat}{\textsc{Llama-2-7B-Chat}}

\newcommand{\random}{Random}
\newcommand{\simdiv}{Sim + Div}
\newcommand{\simi}{Sim}
\newcommand{\ours}{LESS}

\definecolor{c1}{cmyk}{0,0.6175,0.8848,0.1490}
\definecolor{c2}{cmyk}{0.1127,0.6690,0,0.4431}
\definecolor{c3}{cmyk}{0.3081,0,0.7209,0.3255}
\definecolor{c4}{cmyk}{0.6765,0.2017,0,0.0667}
\definecolor{c5}{cmyk}{0,0.8765,0.7099,0.3647}

\newtcbox{\hlprimarytab}{on line, rounded corners, box align=base, colback=c3!10,colframe=white,size=fbox,arc=3pt, before upper=\strut, top=-2pt, bottom=-4pt, left=-2pt, right=-2pt, boxrule=0pt}
\newtcbox{\hlsecondarytab}{on line, box align=base, colback=red!10,colframe=white,size=fbox,arc=3pt, before upper=\strut, top=-2pt, bottom=-4pt, left=-2pt, right=-2pt, boxrule=0pt}

\newtcbox{\flatbox}{on line, box align=base, colback=c4!10,colframe=white,size=fbox,arc=3pt, before upper=\strut, top=-2pt, bottom=-4pt, left=-2pt, right=-2pt, boxrule=0pt}

\newcommand{\dashifted}{\raisebox{0.5\depth}{\tiny$\downarrow$}}
\newcommand{\uashifted}{\raisebox{0.5\depth}{\tiny$\uparrow$}}

\newcommand{\ua}[1]{{\small\hlprimarytab{\uashifted{#1}}}}
\newcommand{\da}[1]{{\small\hlsecondarytab{\dashifted{#1}}}}
\newcommand{\fa}[1]{{\small\flatbox{{#1}}}}

\newcommand{\uag}[1]{{\scriptsize\hlprimarytab{\uashifted{#1}}}}
\newcommand{\dab}[1]{{\scriptsize\hlsecondarytab{\dashifted{#1}}}}

\newcommand*{\TakeFourierOrnament}[1]{{%
\fontencoding{U}\fontfamily{futs}\selectfont\char#1}}
\newcommand*{\danger}{\TakeFourierOrnament{66}}

\definecolor{gg}{HTML}{0F9D58}
\definecolor{rr}{HTML}{DB4437}
\definecolor{bb}{HTML}{4285F4}

\def\dsub{{\gD_{\text{train}}}}
\def\dfull{{\gD}}
\def\dval{{\gD_{\text{val}}}}
\def\dwarmup{{\gD_{\text{warmup}}}}
\newcommand{\valsub}[1]{{\gD_{\text{val}}^{(#1)}}}
\newcommand{\dtest}{\gD_{\text{test}}}
\def\numval{{m}}
\def\dim{{d}}
\newcommand{\grad}[2]{\nabla\ell(#1;#2)}
\newcommand{\loragrad}[2]{\hat\nabla\ell(#1;#2)}
\newcommand{\avggrad}[1]{\bar\nabla\ell(#1)}
\newcommand{\projgrad}[2]{\tilde\nabla\ell(#1;#2)}
\newcommand{\match}[2]{\gM(#1,#2)}
\newcommand{\diverse}[2]{\gK(#1,#2)}
\newcommand{\update}[2]{\Gamma(#1, #2)}
\newcommand{\loraupdate}[2]{\hat\Gamma(#1, #2)}
\newcommand{\approxupdate}[2]{\tilde\Gamma(#1, #2)}
\newcommand{\infadam}{{\textrm{Inf}_{\text{Adam}}}}
\newcommand{\infsgd}{{\textrm{Inf}_{\text{SGD}}}}

\newcommand{\std}[1]{\text{\scriptsize{(#1)}}}


\begin{abstract}
Instruction tuning has unlocked powerful capabilities in large language models (LLMs), using combined datasets to develop general-purpose chatbots.
However, real-world applications often require a specialized suite of skills (e.g., reasoning). The challenge lies in identifying the most relevant data from these extensive datasets to effectively develop specific capabilities, a setting we frame as \textit{targeted instruction tuning}.
We propose LESS, an optimizer-aware and practically efficient algorithm to estimate data influences and perform \textbf{L}ow-rank gradi\textbf{E}nt \textbf{S}imilarity \textbf{S}earch for instruction data selection. 
Crucially, \ours{} adapts existing influence formulations to work with the Adam optimizer and variable-length instruction data.
\ours{} first constructs a highly reusable and transferable \emph{gradient datastore} with low-dimensional gradient features and then selects examples based on their similarity to few-shot examples embodying a specific capability.
Experiments show that training on a  LESS-selected 5\% of the data can often outperform training on the full dataset across diverse downstream tasks.
Furthermore, the selected data is highly transferable: smaller models can be leveraged to select useful data for larger models and models from different families.
Our qualitative analysis shows that our method goes beyond surface form cues to identify data that exemplifies the necessary reasoning skills for the intended downstream application. To facilitate future work, we release code and data at \href{https://github.com/princeton-nlp/LESS}{princeton-nlp/LESS}.
\end{abstract}

\section{Introduction}

Instruction tuning has made large language models (LLMs) adept at following human instructions~\citep{ouyang2022training} as versatile chatbots~\citep{openaichatgptblog, openaigpt4blog, anthropicclaudeblog, googlebardblog}. Recent efforts curating highly diverse and wide-ranging instruction tuning datasets~
\citep[][\emph{inter alia}]{alpaca,wang-etal-2023-self-instruct, mukherjee2023orca, xu2023wizardlm} induce remarkably strong generalization even from a small number of examples~\citep{zhou2023lima}. 
Regardless, it remains an open problem to understand how to best utilize these various datasets.

Many real-world applications call for cultivating a specific suite of capabilities in LLMs (e.g., reasoning skills). 
However, training LLMs with mixed instruction tuning datasets can hinder the development of these specific capabilities. 
For example, \citet{wang2023far} demonstrates that LLMs trained on a mix of instruction tuning datasets exhibit worse performance than those trained on a subset of the data. 
Additionally, considering the broad spectrum of user queries and the multitude of skills required to respond to them, there may not always be enough in-domain data available.
Therefore, we hope to be able to effectively use the general instruction tuning data to improve specific capabilities.
We frame this setting as
\textit{targeted instruction tuning}:

\begin{center}
\vspace{-0.5em} \emph{Given just a handful of examples embodying a specific capability, how can we effectively select relevant fine-tuning data from a large collection of instruction datasets?}\end{center}

\vspace{-0.5em}
We approach this problem by prioritizing training on data that directly minimizes loss on a target task instead of relying on surface form features~\citep{gururangan2020don, xie2023data}.  
Inspired by past works estimating the influence of individual training datapoints with gradient information~\citep{pruthi2020estimating,han2023understanding}, we design an optimizer-aware approach to select such data. 
However, straightforward application of this influence formulation faces several challenges unique to the instruction tuning setting:
(1) LLMs are traditionally fine-tuned with the Adam optimizer~\cite{kingma2015adam} instead of the canonical SGD optimizer; 
(2) using sequence-level gradients of variable-length instruction data can derail the influence estimation; and 
(3) the large number of trainable parameters in LLMs makes the computation and storage of gradient information extremely resource-intensive.

We address these concerns in \ours{}, an algorithm that performs \textbf{L}ow-rank gradi\textbf{E}nt \textbf{S}imilarity \textbf{S}earch to select relevant instruction tuning data for a target application, which exhibits the following properties:
\begin{enumerate}
	\item \textbf{Compatible with instruction tuning with Adam (\textsection\ref{sec:prelims} and \textsection\ref{sec:influence_selection})}: \ours{} adapts the gradient features from classical influence formulations~\citep{pruthi2020estimating} to work with the Adam optimizer and variable-length instruction data. 
 The optimization insights and influence formulation may be of independent interest as well.  
	\item \textbf{Efficient (\textsection\ref{sec:grad_features})}: \ours{} uses LoRA~\cite{hu2021lora} and random projections~\cite{Johnson1984ExtensionsOL} to construct a \textit{gradient datastore} with low-dimensional, easily manipulable gradient features that permit efficient and effective dataset selection. The gradient datastore can be reused for new target tasks.
	\item \textbf{Transferable (\textsection \ref{subsec:main})}: Data selected using small models' gradient features induce strong performance in large models and models from different families, adding to the efficiency of \ours{} (\Cref{tab:main_result}).
	\item \textbf{Interpretable (\textsection\ref{subsec:qualitative})}: Qualitative analysis shows that \ours{} selects data with similar reasoning and skill types as the target task, whereas existing approaches often select data based on surface form cues (e.g., language or topic).
\end{enumerate}

We evaluate our approach on three diverse downstream datasets—\mmlu{}~\cite{hendrycks2020measuring}, \tydiqa{}~\cite{tydiqa}, and \bbh{}~\cite{suzgun2023challenging}—each containing distinct subtasks that effectively simulate targeted instruction tuning scenarios. Results show that \ours{} often selects a small subset of the data ($5\%$) that outperforms training on the full dataset, and the selected subset remains universally effective across model scales and families (\Cref{tab:main_result}).  Comparisons with other data selection methods show that \ours{} is the \emph{only} consistently effective approach, justifying its relatively high computational cost.


\section{Preliminaries: Influence Formulation}\label{sec:prelims}

We restate the approach in~\citet{pruthi2020estimating} of using a first-order approximation of training dynamics to estimate the influence of a training datapoint on held-out data.\footnote{We discuss in~\Cref{sec:related_work} how this formulation of influence is distinct from influence functions~\citep{koh2017understanding}.}

\paragraph{Per-step influence.} Consider a model $\vtheta^t$ at time step $t$ trained on the loss $\ell(\cdot;\vtheta^t)$.
We can write the first-order Taylor expansion of the loss on a validation datapoint $\vz'$ as 
\begin{equation*}
	\ell(\vz';\vtheta^{t+1}) \approx \ell(\vz';\vtheta^t) + \langle\nabla\ell(\vz';\vtheta^t), \vtheta^{t+1}-\vtheta^t\rangle 
\end{equation*}
For ease of exposition, assume that we are training the model with SGD with batch size $1$ and learning rate $\eta_t$.\footnote{We discuss learning rate schedules and batches in~\Cref{sec:nuance}.}
If $\vz$ is the training data at time step $t$, we can write the SGD update as $\vtheta^{t+1} - \vtheta^t = -\eta_t \nabla\ell(\vz;\vtheta^t)$.
Then, the Taylor expansion can be written as
\begin{gather*}
	\resizebox{\linewidth}{!}{$
	\ell(\vz';\vtheta^{t+1}) - \ell(\vz';\vtheta^t) \approx -\eta_t \langle\nabla\ell(\vz;\vtheta^t), \nabla\ell(\vz';\vtheta^t)\rangle $}
\end{gather*}

\paragraph{Trajectory influence.}
The influence of $\vz$ over the entire training run can be measured by aggregating the influence at every training step that uses $\vz$.
Since $\vz$ is used once per epoch, it is natural to express this as a summation over epochs:
\begin{equation}
\infsgd(\vz, \vz') \triangleq \sum_{i=1}^N \bar\eta_i \langle \grad{\vz'}{\vtheta_i}, \grad{\vz}{\vtheta_i}\rangle \label{eq:infsgd}
\end{equation}
where $\bar\eta_i$ is the learning rate used during the $i$th epoch out of $N$ total training epochs and $\vtheta_i$ is the model after the $i$th epoch of training. 

\paragraph{Data selection with influence.} 
While~\citet{pruthi2020estimating} used this insight to identify mislabeled training data, we instead apply this formula to design a data selection strategy. 
In particular, at each time step $t$, selecting $\vz$ to maximize $\langle\nabla\ell(\vz';\vtheta^t), \nabla\ell(\vz;\vtheta^t)\rangle$ will drive a larger decrease in the loss on the validation point $\vz'$. 
However, when computing $\infsgd$ across several epochs, we note that the model checkpoints $\{\vtheta_i\}$ after the first epoch will depend on the dataset selected for training.
This causes the data selection problem to become circular, and we empirically circumvent this problem with a short \textit{warmup training} run on a randomly selected $\mathcal{D}_\mathrm{warmup} \subset \gD$ for $N=4$ epochs (see \textsection\ref{sec:grad_features}).
Overall, this data selection strategy is especially useful in the transfer learning setting, because it does not require any specific relationship between $\vz'$ and $\vz$.
The next two sections describe how we adapt this basic approach to operate efficiently and effectively with instruction tuning.

\section{\ours{}: Estimating the Influence of Instructions}
\label{sec:influence_selection}
Here, we describe how \ours{} adapts the influence formulation in~\Cref{eq:infsgd} to select instructions that effectively induce a target capability. 
We consider two main conceptual challenges here: (1) past works described in~\Cref{sec:prelims} focus on the SGD optimizer, but LLMs are traditionally tuned using Adam (\textsection\ref{subsec:adam_ex}), and (2) instruction tuning exhibits abnormal gradients that cause the prior formulation to heavily favor selecting shorter instructions, thereby hurting performance  (\textsection\ref{subsec:inst_grad}).

\paragraph{Notation.}
We follow the standard autoregressive setting for instruction tuning, where a datapoint $\vz$ consists of an instruction $\vs$ and a completion $\vc$.\footnote{Some datasets contain multi-turn instructions, but we denote them as single turn instructions for ease of exposition.} 
For each data point, the loss $\ell(\vz;\vtheta)$ is the cross entropy between the model distribution $p(\cdot\vert\vs)$ and the true label $\vc$. When $\vc$ contains multiple tokens, we use an average of the token-wise cross-entropies in the completion as the loss of $\vz$. 
We use $\vtheta^t$ to indicate the model at the $t$-th step, and we use $\vtheta_i$ to indicate the model after the $i$th epoch of training. 

\subsection{Extension to Adam}
\label{subsec:adam_ex}
The formulation in~\Cref{eq:infsgd} is unique to optimizing models with SGD. However, instruction tuning is usually performed using the Adam optimizer~\citep{kingma2015adam}.\footnote{\Cref{app:adam_sgd_ft} demonstrates instruction tuning with SGD significantly hurts performance.} In this case, the parameter update at a given step is: 
\begin{gather*}
	\vtheta^{t+1} - \vtheta^t = -\eta_t \update{\vz}{\vtheta^t} \\
	\update{\vz}{\vtheta^t} \triangleq \frac{\vm^{t+1}}{\sqrt{\vv^{t+1} + \epsilon}} \\
	\vm^{t+1} = (\beta_1\vm^t + (1-\beta_1)\grad{\vz}{\vtheta^t}) / (1-\beta_1^t) \\
	\vv^{t+1} = (\beta_2\vv^t + (1-\beta_2)\grad{\vz}{\vtheta^t}^2) / (1-\beta_2^t) 
\end{gather*}
where all operations are performed elementwise, with $\beta_1$ and $\beta_2$ as the hyperparameters for the first and second moments, respectively, and $\epsilon$ as a small constant.
Then, the first-order expansion for the Adam dynamics suggests we should choose $\vz$ to maximize $\langle \grad{\vz'}{\vtheta^t}, \update{\vz}{\vtheta^t}\rangle$.
Note that extending the data selection strategy to Adam exacerbates the aforementioned circularity of the procedure, because computing $\update{\vz}{\vtheta}$ requires accessing the $\vm$ and $\vv$ terms, which are determined by prior training gradients.
As before, we obtain these from the warmup training (\textsection\ref{sec:grad_features}).\footnote{We ablate against using $\textrm{sign}(\grad{\cdot}{\vtheta_i})$ as an approximation for $\update{\vz}{\vtheta_i}$, as suggested in past theoretical works~\citep{malladi2022sdes,malladi2023kernel}. 
See~\Cref{app:adam_sgd}.}

\subsection{Addressing Abnormalities in Sequence-Level Gradients}
\label{subsec:inst_grad}
A faithful application of \Cref{sec:prelims} to the instruction tuning setting would consider that the gradient of an example $\vz$ is the average of the gradients of each token in the completion.
In particular, we would choose which tokens to train on within each sequence in order to best reduce the validation loss, and notably, optimizing the inner product at a fine-grained level may lead to better results.
However, estimating individual token influences is extremely expensive because it requires computing the gradient of a vector of token losses with respect to the model parameters, so we instead perform data selection at the sequence level.

Because the example gradient is the average of many token gradients, the gradient norm of the sequence $\|\grad{\vz}{\vtheta^t}\|$ is negatively correlated with the length of the completion (\Cref{fig:grads_length,fig:grads_length_adam}). 
This is a well-known issue for influence formulations that compute averaged token gradients for sequences of varying lengths. 
We discuss this observation further in~\Cref{sub:grad_norm} but defer an in-depth study to future work.
This phenomenon causes the influence in~\Cref{eq:infsgd} to heavily upweight shorter training sequences, which in turn leads to markedly worse performance (see~\Cref{tab:normalized_grads}).
To mitigate this problem, we normalize the gradient features in \ours{} and use the cosine similarity instead of the dot product to estimate influences.
\begin{definition}[Adam Influence]\label{def:influence}
	Suppose the model is trained for $N$ epochs, where $\bar\eta_i$ is the average learning rate in the $i$th epoch and $\vtheta_i$ is the model checkpoint after the $i$th epoch.
	We define the influence of a training datapoint $\vz$ on a validation datapoint $\vz'$ when training with Adam as 
	\begin{gather*}
	\infadam(\vz, \vz') \triangleq \sum_{i=1}^N \bar\eta_i \cos(\grad{\vz'}{\vtheta_i}, \update{\vz}{\vtheta_i}) \label{eq:infadam}
	\end{gather*}
	\end{definition}
 where $\cos$ computes the cosine similarity of the two vectors.

\section{\ours{}: Estimating Influences Efficiently}\label{sec:efficient_influences}

\begin{figure*}
	\centering
	\includegraphics[width=\linewidth]{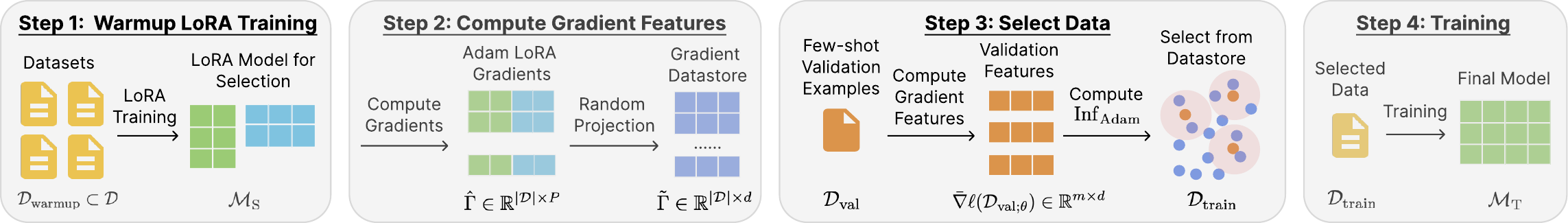}
	\caption{Illustration of \ours{}. In step 1, we train a selection model $\mathcal{M}_S$ with LoRA for a warmup period with a small subset of data $\mathcal{D}_{\mathrm{warmup}}\subset \mathcal{D}$. In step 2, we compute the Adam LoRA gradient features $\Gamma\in\sR^{|\dfull|\times P}$ for each candidate datapoint  and save them in a gradient datastore. In step 3, for any task with few-shot examples $\dval$ (comprising of $m$ subtasks), we compute the gradient features for each validation subtask and select the subset $\mathcal{D}_{\text{train}}$ with the top $5\%$ training examples ranked by $\infadam$. Step 4 is the final training stage with the selected data on a target model $\mathcal{M}_T$, which can be trained with either LoRA or full finetuning. Steps 1 and 2 are offline and only need to be computed once per candidate training set $\dfull$. }
	\label{fig:method}
\end{figure*}
We have established how to estimate the influence using model gradients, but given the sheer size of LLMs, computing and storing model gradients is still prohibitively expensive.
In this section, we describe the design choices that make \ours{} computationally efficient, and illustrate the computing steps in~\Cref{fig:method}. 

We use LoRA~\citep{hu2021lora} to efficiently perform the warmup training.
Then, we construct a \emph{gradient datastore} of projected low-dimensional gradient features~\citep{Johnson1984ExtensionsOL, park2023trak} that can be reused for different target tasks. 
Finally, we detail how our data selection algorithm uses the datastore to construct a training dataset that accommodates substructures (e.g., individual reasoning skills) in the target validation set.

\paragraph{Setting.}
Our goal is to select a subset $\dsub$ of a large, diverse instruction tuning dataset $\dfull$ such that training a model on $\dsub$ achieves lower loss on a target validation dataset $\dval$, and therefore performs well on the test data.  
$\dval$ can contain several subtasks $\valsub{1}, ..., \valsub{m}$ (e.g., question answering for multiple languages).\footnote{This is case for datasets we study in this paper, but our method also applies to datasets with no explicit substructures (i.e., $m=1$).}
The validation set is fixed and only contain a few examples per subtask (e.g., 1 example per language in \tydiqa{}). 
We use a held-out test set $\dtest$ to evaluate the performance of the model trained on $\dsub$.
We use $\gM_S$ to denote the model parametrized by $\vtheta$ used for scoring and selecting data, and we use $\gM_T$ to denote the model trained on the selected data.
The selection model $\gM_S$ need not be the same as the target model $\gM_T$; In this case, we refer to the setting as \ours{}-T.

\subsection{Efficient Gradient Feature Computation}
\label{sec:grad_features}
Computing \Cref{def:influence} requires taking the inner product of two vectors with a size of trainable parameters in the model.
Directly using such high-dimensional gradient vectors as features for dataset selection is very computationally expensive, so we apply two techniques to construct meaningful low-dimensional gradient features: parameter efficient fine-tuning via LoRA~\citep{hu2021lora} and random projections~\citep{Johnson1984ExtensionsOL, park2023trak}.
We apply these techniques to $\grad{\vz'}{\vtheta}$ for validation datapoints $\vz'$ and to $\update{\vz}{\vtheta}$ for training datapoints $\vz$.
To enable rapid transfer to new target validation tasks, we create a datastore with these low-dimensional features for candidate training datapoints.

\paragraph{Step 1: Warmup training with LoRA.}
We use LoRA~\citep{hu2021lora} to reduce the number of trainable parameters and accelerate the inner products in~\Cref{def:influence}.
LoRA freezes the pre-trained weights and adds a low-rank adaptor to linear layers throughout the network.
We use LoRA to instruction tune a pre-trained base model (e.g., \llamasmall{}) on a random subset $\mathcal{D}_\mathrm{warmup} \subset \gD$ for $N$ epochs (we only use 5\% of the training data in practice, see \textsection\ref{subsec:setup}), checkpointing the model after each epoch to store $\{\vtheta_i\}_{i=1}^N$.
The gradient when training with LoRA, denoted $\loragrad{\cdot}{\vtheta}\in\sR^P$, is much lower dimensional than the model itself; for example,  in \llamasmall{}, $\loragrad{\cdot}{\vtheta}$ is less than $2\%$ the size of $\vtheta$. 
We use $\loragrad{\cdot}{\vtheta}$ to compute the Adam update and denote it as $\loraupdate{\cdot}{\vtheta}$. 
This initial warmup training is motivated conceptually in~\textsection \ref{subsec:adam_ex}, and empirical results in \textsection \ref{subsec:efficiency} demonstrate that omitting it yields suboptimal results. 


\paragraph{Step 2: Projecting the gradients.}
To further reduce the feature dimensionality, we apply a random projection to the LoRA gradients. 
The Johnson-Lindenstrauss Lemma~\citep{Johnson1984ExtensionsOL} asserts that such projections often preserve the inner products in~\Cref{def:influence}, thereby ensuring these low-dimensional gradient features are still useful for dataset selection. 
For a given validation datapoint $\vz'$ and model checkpoint $\vtheta_i$, we can compute a $\dim$-dimensional projection of the LoRA gradient $\projgrad{\vz'}{\vtheta_i} = \Pi^\top \loragrad{\vz'}{\vtheta_i}$, with each entry of $\Pi\in\sR^{P\times\dim}$ drawn from a Rademacher distribution (i.e., $\Pi_{ij}\sim\gU(\{-1,1\})$). 
For training datapoints $\vz$, we compute $\approxupdate{\vz}{\cdot} = \Pi^\top\loraupdate{\vz}{\cdot}$.  

We use the memory-efficient online implementation of random projections from~\citet{park2023trak} to compute and apply $\Pi$.
In practice, we choose $\dim=8192$.

\subsection{Data Selection Algorithm}\label{sec:alg}
The data selection strategy (see step 3 in \Cref{fig:method}) adapts the approach in~\Cref{sec:prelims} to efficiently operate on subtasks of the validation set (e.g., individual tasks within BBH).
For each subtask $\valsub{j}$, we compute its average gradient feature for every model checkpoint $\vtheta_1,...,\vtheta_N$: 
\begin{equation*}
	\avggrad{\valsub{j};\vtheta_i} = \frac{1}{|\valsub{j}|}\sum_{\vz'\in\valsub{j}} \projgrad{\vz'}{\vtheta_i}.
\end{equation*}
As in~\Cref{def:influence}, we then aggregate the scores of how close the given datapoint is to each validation subtask throughout training. 
For example, we can adapt~\Cref{def:influence} to write
\begin{equation}
	\infadam(\vz, \valsub{j}) = \sum_{i=1}^N \bar\eta_i \frac{\langle \avggrad{\valsub{j}; \vtheta_i}, \approxupdate{\vz}{\vtheta_i} \rangle }{\|\avggrad{\valsub{j}; \vtheta_i}\|\|\approxupdate{\vz}{\vtheta_i}\|}.
 \label{eq:matching}	
\end{equation}
We select training datapoints that can improve performance on any one of the validation subtasks.
Following the logic in~\textsection\ref{sec:prelims}, we compute the score for $\vz$ as the maximum across all subtasks: $\max_j \infadam(\vz, \valsub{j})$.
We select the highest scoring examples to construct $\dsub$.\footnote{We ablate against the analogous form of the algorithm using $\infsgd$ in~\Cref{eq:infsgd} for data selection. See \Cref{app:adam_sgd}.} After selection, we use the selected subset $\dsub$ to train the target model $\gM_T$.

\section{Experiments}\label{sec:experiments}
We introduce the experimental setup (\textsection\ref{subsec:setup}) and relevant baselines (\textsection\ref{subsec:baselines}), and then we present the main results (\textsection\ref{subsec:main}). 

\subsection{Experimental Setup}
\label{subsec:setup}
\paragraph{Training datasets.} We follow \cite{wang2023far} and use the following instruction tuning datasets: (1) datasets created from existing ones such as \flan~\cite{longpre2023flan} and \cott~\cite{wei2022chain}; (2) open-ended generation datasets with human-written answers including \dolly~\cite{DatabricksBlog2023DollyV2} and \oasst~\cite{kopf2023openassistant}. The datasets, comprising approximately 270K data points, vary widely in their format and underlying reasoning tasks, and do not contain any obvious in-domain data for the target queries.  
\Cref{app:train_data} contains more dataset details.

\begin{table}[h]
    \caption{Statistics of evaluation datasets. The selection of evaluation tasks cover different kinds of answer types.}
    \resizebox{0.49\textwidth}{!}{%
    \begin{tabular}{lrrrrl}
    \toprule
    \textbf{Dataset}                & \textbf{\# Shot} & \textbf{\# Tasks} & $\vert \boldsymbol{\mathcal{D}}_{\boldsymbol{\mathrm{val}}} \vert$ & $\vert \boldsymbol{\mathcal{D}}_{\boldsymbol{\mathrm{test}}} \vert$ & \textbf{Answer Type} \\ \midrule
    \mmlu   &5 & 57   & 285    & 18,721        & Letter options \\ 
    \tydiqa & 1 & 9   & 9      & 1,713         & Span \\
    \bbh    & 3 & 23   & 69   & 920          & COT and answer \\ \bottomrule 
    \end{tabular}}
    \label{tab:eval_dataset}
\end{table}

\paragraph{Evaluation datasets.} We evaluate our method on \mmlu~\cite{hendrycks2020measuring}, \tydiqa~\cite{tydiqa} and \bbh~\cite{suzgun2023challenging}.~\mmlu{} consists of multiple-choice questions spanning 57 tasks including elementary mathematics, US history, computer science, law, and more.~\tydiqa{} is a multilingual question-answer dataset containing 9 typologically diverse languages. Given a question and a related passage, the task requires extrating the answer from the passage.~\bbh{} is a collection of 27 challenging tasks from BIG-Bench selected to evaluate reasoning capabilities.
Table~\ref{tab:eval_dataset} contains more details about these tasks. 
Each dataset includes multiple subtasks, and each subtask comes with few-shot examples. 
These examples are used as $\dval$ for data selection (\textsection\ref{sec:alg}) and as few-shot in-context learning demonstrations in evaluation. \Cref{app:eval} contains more dataset details.

\paragraph{Models for data selection and training.} 
We test \ours{} with three base models: \llamasmall{}~\citep{touvron2023llama2}, \llamalarge{}~\citep{touvron2023llama2}, and \mistral{}~\citep{jiang2023mistral}.
In the transfer setting \ours-T, we select using \llamasmall{} as $\gM_S$ and train either \llamalarge{} or \mistral{} as the target model $\gM_T$.  
The warmup training and the final model training are both conducted with LoRA~\cite{hu2021lora}. 
We report the average performance and standard deviation across three random seeds.
\Cref{app:training_setup} contains more training details.  

\paragraph{Default setting.} 
\ours{} performs warmup training on a randomly selected 5\% of the complete dataset $\mathcal{D}_\mathrm{warmup}$ for $N=4$ epochs and computes $8192$-dimensional gradient features on the data $\dfull$ (\textsection\ref{sec:grad_features}). 
For each target task, we use these features to score datapoints by their influences (\Cref{def:influence}) and select the highest scored 5\% of $\dfull$ to construct $\dsub$.
We train the target model $\gM_T$ on this selected data $\dsub$.

\subsection{Baselines}
\label{subsec:baselines}
We compare \ours{} with a several baselines~(see more details in \Cref{app:baselines}). The simplest baseline is \textbf{random selection}, where we randomly sample data from the training set for instruction tuning. We also compare to \textbf{BM25}~\citep{robertson2009probabilistic}, which featurizes examples by their word frequency statistics (i.e., TF-IDF) to rank the training instances, and select the top $k$ data points with the highest scores to construct $\dsub$.
Another baseline is \textbf{DSIR}~\citep{xie2023data}, which uses $n$-gram features to weight candidate training data $\dfull$. We resample $k$ data points accoding to the importance weights. 
We also compare to \textbf{RDS} (Representation-based Data Selection)~\citep{zhang2018unreasonable, hanawa2020evaluation}, which uses model's hidden representations as features for data selection. For fair comparison, we compute the similarity score using~\Cref{eq:matching} but replace the gradient features with the final layer representations of the last token of each sequence. These features are $2048$-dimensional in \llamasmall{}. Another intuitive baseline we consider is to directly use the validation data $\dval$ for training, but this drastically degrades performance (\Cref{app:valid}).

\subsection{Main results}

\begin{table*}[t]
    \caption{Results of \ours{} (selected with target model's gradient store) and \ours{}-T (selected with \llamasmall's gradient store) on \llamasmall{}, \llamalarge{} and \mistral{}. Full denote full dataset, and otherwise we select 5\% of the data with random selection (Rand.), \ours-T and \ours{}. \textbf{Bold} numbers denotes the best performing selected subset. \underline{Underlined} numbers denote that the selected subset outperforms the full dataset. Numbers in the parentheses are standard deviations.}
    \resizebox{\textwidth}{!}{%
    \setlength{\tabcolsep}{3pt}
    \begin{tabular}{lcccc|cccc|cccc} \toprule
 & \multicolumn{4}{c}{\textbf{\mmlu}}             & \multicolumn{4}{c}{\textbf{\tydiqa}}           & \multicolumn{4}{c}{\textbf{\bbh}}              \\
 \cmidrule(lr){2-5} \cmidrule(lr){6-9} \cmidrule(lr){10-13}
    & \textbf{Full} & \textbf{Rand.} & \textbf{\ours{}-T} & \textbf{\ours}  & \textbf{Full}       & \textbf{Rand.} & \textbf{\ours{}-T} & \textbf{\ours} & \textbf{Full} & \textbf{Rand.} & \textbf{\ours{}-T} & \textbf{\ours}         \\
    \textbf{Data percentage} & \text{\small (100\%)} & \text{\small (5\%)} & \text{\small (5\%)} & \text{\small (5\%)} & \text{\small (100\%)} & \text{\small (5\%)} & \text{\small (5\%)} & \text{\small (5\%)} & \text{\small (100\%)} & \text{\small (5\%)} & \text{\small (5\%)} & \text{\small (5\%)}  \\
    \cmidrule(r){1-1}     \cmidrule(lr){2-2} \cmidrule(lr){3-5} \cmidrule(lr){6-6} \cmidrule(lr){7-9} \cmidrule(lr){10-10} \cmidrule(lr){11-13}
    \textbf{\llamasmall} & \textit{51.6} & 46.5 \std{0.5} & - & \textbf{50.2} \std{0.5} & \textit{54.0} & 52.7 \std{0.4} & - & \underline{\textbf{56.2}} \std{0.7} & \textit{43.2} & 38.9 \std{0.5} & - & \textbf{41.5} \std{0.6} \\ 
    \textbf{\llamalarge} & \textit{54.5} & 53.4 \std{0.1} & \underline{\textbf{54.6}} \std{0.3} & 54.0 \std{0.7} & \textit{54.3} & 53.0 \std{1.3} & \underline{\textbf{57.5}} \std{0.8} & \underline{54.6} \std{0.3} & \textit{50.8} & 47.0 \std{1.6} & 49.9 \std{0.5} & \textbf{50.6} \std{0.6} \\ 
    \textbf{\mistral} & \textit{60.4} & 60.0 \text{\scriptsize{(0.1)}} & \underline{60.6} \text{\scriptsize{(0.3)}} & \underline{\textbf{61.8}} \text{\scriptsize{(0.4)}} & \textit{57.7} & 56.9 \text{\scriptsize{(0.2)}} & \underline{\textbf{61.7}} \text{\scriptsize{(1.7)}} & \underline{60.3} \text{\scriptsize{(2.4)}} & \textit{53.0} & 54.5 \text{\scriptsize{(0.1)}} & \underline{\textbf{56.0}} \text{\scriptsize{(0.8)}} & \underline{\textbf{56.0}} \text{\scriptsize{(1.0)}} \\
    \bottomrule
    \end{tabular}}
 
    \label{tab:main_result}
\end{table*}

\begin{table}[h]
        \caption{Comparison of \ours{} with BM25, DSIR, and RDS methods when training with top 5\% of the data on \llamasmall. $\Delta$ denotes the performance gain of \ours{} over the strongest baseline. }
    \centering
    \resizebox{\columnwidth}{!}{%
    \setlength{\tabcolsep}{2pt}
    \begin{tabular}{lcccccc}
    \toprule
    & \textbf{Rand.} & \textbf{BM25} & \textbf{DSIR} & \textbf{RDS} & \textbf{\ours{}} & $\Delta$ \\ \cmidrule(r){2-7}
    \mmlu   & 46.5 \std{0.5} & 47.6 & 46.1 \std{0.3} & 45.0 \std{1.0} & \textbf{50.2} \std{0.5} & \ua{2.6} \\ 
    \tydiqa & 52.7 \std{0.4} & 52.7 & 44.5 \std{1.7} & 46.8 \std{1.3} & \textbf{56.2} \std{0.7} & \ua{3.5}  \\ 
    \bbh    & 38.9 \std{0.5} & 39.8 & 36.8 \std{0.1} & 36.7 \std{1.3} & \textbf{41.5} \std{0.6} & \ua{1.7}  \\ \bottomrule
    \end{tabular}}
    \label{tab:baselines}
\end{table}
\label{subsec:main}
We present main results of \ours{} across different models in \Cref{tab:main_result}, and comparison with baseline approaches in \Cref{tab:baselines}. We summarize the findings below.

\paragraph{\ours{} is effective across different models.} \Cref{tab:main_result} shows that \ours{} consistently outperforms random selection by $2$ to $5$ points across all models and evaluation datasets, demonstrating that our data selection approach selects useful data for fine-tuning.

\paragraph{5\% of the data often outperforms the full dataset.} In \Cref{tab:main_result}, we compare training with 5\% of the data selected by \ours{} and training with the full dataset. Surprisingly, we discover that in most cases, training with just 5\% of the data yields better results than using the entire dataset. We hypothesize that the full dataset contains irrelevant or detrimental data points for specific target tasks. This effect is more evident when using stronger base models (e.g., \llamalarge~and \mistral).~\footnote{
    We note that there may still be useful data in the discarded $95\%$, and we leave exploring the optimal threshold to future work.}

\paragraph{Data selected using a small model boosts performance for larger and different models.} \Cref{tab:main_result} evaluates several \ours-T settings.
We use \llamasmall{} as the selection model $\gM_S$ and train either \llamalarge{} (i.e., a larger model) and \mistral{} (i.e., a different model family) as the target model $\gM_T$.
\ours{}-T outperforms random selection and has a relatively small gap compared to \ours{} (i.e., directly using \llamalarge{} or \mistral{} as the selection model $\gM_S$). 
The strong performance of \ours{}-T  indicates that \ours{} selects generalizably useful data, so data selection costs can be further amortized across different models.
Conceptually, our derivation in~\Cref{sec:prelims} states that the dataset selected using one model induces strong performance in another one when the inner products of the gradients of different examples are roughly equal for $\gM_S$ and $\gM_T$. 
Our experiments support that this finding is true, thereby adding to the growing body of work that small models can effectively select data for other models during pre-training~\citep{xie2023doremi,engstrom2024dsdm} and in-context learning~\citep{wang2023large}.

\paragraph{\ours{} is the only consistently effective approach compared to others.}
In \Cref{tab:baselines}, we compare \ours{} to several baselines. We observe that conventional approaches relying on word frequency statistics (e.g., BM25), $n$-gram features (e.g., DSIR), or representations (e.g., RDS) show minimal improvement over random selection, and \ours{} consistently outperforms the strongest baseline. 
We acknowledge that the baseline methods are less computationally expensive. 
Nonetheless, our findings suggest that instruction data calls for a refined definition of similarity between data points (i.e., the influence formulation in~\Cref{def:influence}). 
Although the baseline approaches typically select data that is topically or semantically similar to the query and completion, qualitative analysis (\textsection\ref{subsec:qualitative}) indicates that \ours{} is better at selecting data that require the same kind of reasoning process (i.e., how the instruction is executed) as the target task $\dval$.

\begin{table*}[ht]
    \caption{Asymptotic complexity,  wall-clock runtime (measured as \textbf{single} A100 GPU hours) and storage cost associated with each step in \ours. Gradient computation is the most costly step, followed by the warmup LoRA training stage, but this expense is incurred only once. The actual data selection process requires minimal computation.}
    \centering
    \resizebox{0.9\textwidth}{!}{%
    \begin{tabular}{@{}lcc|cc|cc@{}}
    \toprule
            & \multicolumn{2}{c}{\textbf{Warmup LoRA Training}}      & \multicolumn{2}{c}{\textbf{Gradient Features Computation}}        &\multicolumn{2}{c}{\textbf{Data Selection}}        \\ \midrule
            & \textbf{Complexity} & \textbf{Actual} & \textbf{Complexity} & \textbf{Actual} & \textbf{Complexity} & \textbf{Actual} \\ 
            \cmidrule(lr){2-3} \cmidrule(lr){4-5} \cmidrule(lr){6-7} 
    \textbf{Compute} & $\mathcal{O}(\vert \mathcal{D}_{\mathrm{warmup}} \vert \cdot N)$ & $6$ Hours & $\mathcal{O}(\vert \dfull \vert \cdot N)$ & 48 Hours & $\mathcal{O}(\vert \dfull \vert \cdot \vert \dval \vert \cdot d)$        & $< 1$ Min  \\ 
    \textbf{Storage} &  - & -                                 & $\mathcal{O}(\vert \dfull \vert \cdot N \cdot \dim)$ & 17.7 GB       & - & -  \\ 
    \bottomrule
    \end{tabular}}
    \label{tab:efficiency}
\end{table*}

\paragraph{Additional models and tasks.} We also perform experiments on more model families and tasks. 
(1) We conduct experiments with Pythia models~\citep{biderman2023pythia} across different scales: 14M, 410M, 1B, 6.9B, and 12B. 
Our findings indicate that LESS consistently outperforms random selection across different model scales. 
Surprisingly, models as small as Pythia-14M can still effectively select data for larger Pythia models despite not being able to solve the task themselves. 
Please find more details in \Cref{subsec:pythia}.  
(2) In experiments on GSM8K~\citep{cobbe2021training} and TruthfulQA~\citep{lin2021truthfulqa}, \ours{} significantly outperforms random selection, and on TruthfulQA, using just 5\% of the data even outperforms using the full dataset. 
Please find more details in \Cref{subsec:gsm8k_truthfulqa}.
\section{Analysis}\label{sec:analysis}
\looseness-1
We analyze \ours{} in two ways.
First, we detail the computational cost of \ours{} and explore how different design choices affect performance (\textsection\ref{subsec:efficiency}).
Second, we present a qualitative analysis showing that \ours{} selects data that aligns with the reasoning capabilities needed for the target task (\textsection\ref{subsec:qualitative}).
\begin{table}[t]
        \caption{Performance of using vanilla gradients from off-the-shelf models (e.g., \llamasmall{} and \llamasmallchat{}), and Adam gradients from LoRA models (trained with a random set of 5\%, 25\% and 100\% of the full dataset for warmup) for data selection. Unk. denotes that the instruction tuning data is unknown.}
    \centering
    \setlength{\tabcolsep}{3pt}
    \resizebox{\columnwidth}{!}{%
    \begin{tabular}{@{}lccccc@{}}
    \toprule
    & \multicolumn{2}{c}{\textbf{\llamasmall}} & \multicolumn{3}{c}{\textbf{LoRA Models}} \\
     \cmidrule(lr){2-3} \cmidrule(lr){4-6}
     & \textbf{Base (0\%)}     & \textbf{Chat} \text{\small (Unk.)}      & \textbf{5\%} \text{\small (default)}      & \textbf{25\%}     & \textbf{100\%}     \\
    \cmidrule(r){1-1} \cmidrule(lr){2-3} \cmidrule(lr){4-6}
    \textbf{\mmlu} & 46.7    & 47.9         & 50.2     & 51.3     & 51.6      \\
    \textbf{\tydiqa} & 52.1    & 52.2         & 56.2     & 57.0     & 57.9      \\
    \textbf{\bbh} & 39.8    & 38.6         & 41.5     & 41.5     & 41.9      \\ 
    \cmidrule(r){1-1} \cmidrule(lr){2-6}
    \textbf{Avg.} & 46.2    & 46.2         & 49.3     & 49.9     & \textbf{50.5}      \\ \bottomrule
    \end{tabular}}
    \label{tab:warmup_data}
\end{table}

\subsection{Computational Complexity}
\label{subsec:efficiency}
\autoref{tab:efficiency} describes the asymptotic complexity, wall-clock runtime, and the storage cost required for each step of \ours{}. The reported wall-clock time is measured in \textbf{single} A100 (80GB) GPU hours.
The gradient feature computation is the most expensive step, and the cost scales linearly with candidate dataset size $|\dfull|$, number of checkpoints $N$, and gradient dimension $\dim$. 
Storing the resulting gradient datastore consumes memory that scales linearly with $\dim$. 
Warmup training is also relatively expensive, and the complexity scales with $|\dwarmup|$ and $N$. 
However, the expenses incurred during these two stages are one-time costs that can be amortized over many target tasks.

In the following sections, we explore how our choices for $|\mathcal{D}_{\mathrm{warmup}}|$, $N$, and $\dim$ impact the performance of \ours. 
In summary, increasing $\dwarmup$ and $N$ improves the accuracy of the influence estimation (\Cref{def:influence}), but doing so also increases the cost of the warmup training step. 
Similarly, increasing $\dim$ improves the utility of the projected gradient features~\citep{Johnson1984ExtensionsOL}, but doing so also scales the memory and compute required for \ours.

\paragraph{Warmup training is crucial.} 
The warmup training step tunes a model on $\dwarmup$, which is usually a random 5\% of the data $\dfull$.
We investigate: (1) if this step is necessary, and (2) if using more data further enhances performance. 
For (1), we use pre-trained \llamasmall{} and \llamasmallchat{} as selection models to create a gradient data store for selecting data.\footnote{Note that we compute $\infsgd$ instead of $\infadam$, since the pre-training optimization states (i.e., $\Gamma$) is not available.} \autoref{tab:warmup_data} demonstrates that using pre-trained models for data selection significantly hurts the performance of \ours{}.
This observation may stem from the input distribution shift, or more generally, nuances in the training dynamics (see~\Cref{sec:kernel}).
We leave detailed investigations to future work. 
For (2), we vary the size of $\dwarmup$ to be 5\% (the default), 25\%, and the entire dataset and compute $\infadam$. 
Increasing $|\dwarmup |$ improves performance, confirming our hypothesis that estimating $\infadam$ (\Cref{eq:infadam}) more accurately is helpful.
Therefore, the warmup phase is crucial, but it does not need to be too long for \ours{} to yield strong performance.

\begin{table}[t]
    \caption{Number of checkpoints ($N$) used for select data with \ours{}. Using fewer checkpoints still outperforms random selection but is less effective.}
    \centering
    \setlength{\tabcolsep}{4pt}
    \resizebox{0.9\columnwidth}{!}{%
    \begin{tabular}{@{}lcccc@{}} \toprule
                  & \textbf{\mmlu} & \textbf{\tydiqa} & \textbf{\bbh} & \textbf{Avg.} \\ \cmidrule(r){1-1} \cmidrule(lr){2-5} 
    Random       & 46.5 \std{0.5} & 52.7 \std{0.4} & 38.9 \std{0.5} & 46.0 \\
    $N=1$           & 48.2 \std{0.4} & 54.9 \std{0.4}   & 40.2 \std{0.2} & 47.8 \\
    \cmidrule(r){1-1} \cmidrule(lr){2-5} 
    $N=4$ (default) & 50.2 \std{0.5} & 56.2 \std{0.7}  & 41.5 \std{0.6} & \textbf{49.3} \\ \bottomrule
    \end{tabular}}
    \label{tab:number_ckpts}
    
\end{table}
\paragraph{More checkpoints, better performance.} We investigate if using fewer checkpoints from the warmup training stage (i.e., summing over fewer gradient features in~\Cref{def:influence}) affects the performance of \ours{}. 
\Cref{tab:number_ckpts} shows that using only one LoRA checkpoint outperforms random selection but underperforms using four checkpoints.
We speculate that this is because the gradient features change substantially at the start of instruction tuning (see~\Cref{sec:kernel}).

\begin{figure}[t]
    \begin{minipage}{0.48\textwidth}
        \centering
        \includegraphics[width=\linewidth]{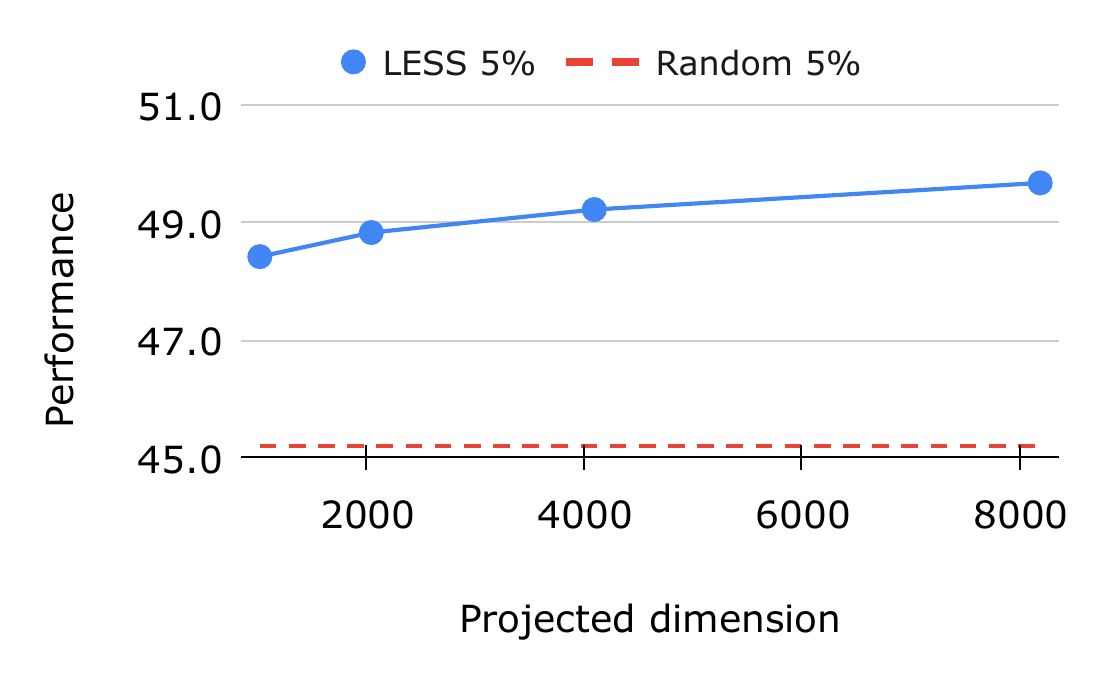}
        \caption{Average performance of \ours{} on three datasets when using different projected dimension (1024, 2048, 4096 and 8192). We select 5\% of the data for instruction tuning. Please refer to \Cref{app:dimension} for task-specific results. }    
        \label{fig:projected_dimension}    
    \end{minipage}
\end{figure}

\paragraph{A small projection dimension is sufficient.} We study how choosing different gradient projection dimensions $d$, specifically 1024, 2048, 4096, and 8192 (default), impacts the performance of \ours. 
\Cref{fig:projected_dimension} shows that all choices of $d$ improve performance over random selection, and as $d$ increases, \ours{} yields a stronger performance at a higher computational cost (see \Cref{tab:efficiency}).

\subsection{Qualitative Analysis}
\label{subsec:qualitative}
We manually compare the data selected by \ours{} and baseline approaches. 
\Cref{tab:tydiqa_example} shows top examples selected by BM25, RDS, and \ours{}, given a single Bengali question-answering example from the \tydiqa{} dataset. 
BM25 and RDS simply match the few-shot data $\dval$ and select Bengali examples from different tasks (mask recovery and hate speech detection, respectively).\footnote{The top 10 examples selected by BM25 and RDS are of Bengali translation, Bengali sentiment classification, Bengali mask prediction, etc.} 
In contrast, \ours{} selects an English example from an open-book question-answering task similar to \tydiqa{}. 
This example demonstrates that \ours{} selects examples with a similar underlying reasoning type without relying heavily on surface textual or linguistic cues. 
Such examples enable effective zero-shot transfer to the Bengali language, despite being in a different language~\citep{pmlr-v119-hu20b}. 
Furthermore, we observe that \ours{} chooses more examples from the \dolly{} dataset for \tydiqa{}, likely because \dolly{} contains many open-book question-answering examples. 
Our findings suggest that \ours{} may be used to improve interpretability of large, general-purpose instruction tuning datasets.
\Cref{app:qualitative} contains more analysis details.

\section{Related Work}\label{sec:related_work}

\paragraph{Curating high-quality instruction tuning data.}
Using high-quality instruction tuning data can dramatically improve base LLMs. 
Instruction tuning data is generally of two types: (1) task-based datasets curated from traditional NLP tasks~\citep{wang2022super,sanh2022multitask,wei2022finetuned,longpre2023flan}, and (2) open-ended instruction following datasets, broadly covering a wide range of topics~\citep{alpaca,DatabricksBlog2023DollyV2,kopf2023openassistant,xu2023wizardlm, mukherjee2023orca,zhou2023lima,ding2023enhancing}. 
Increasing data quality and diversity instead of quantity has been shown to more effectively induce instruction following abilities~\citep{cao2023instruction,chen2023alpagasus,bukharin2023data,du2023mods, liu2023makes, li2023one}. Moreover, ~\citep{han2023context, lin2023unlocking} have demonstrated that in-context learning alone can significantly advance instruction following.
Our work adds to this narrative by selecting high-quality, relevant data from a large pool of available datasets to induce a particular capability in the model.

\paragraph{Coresets and gradient-based data selection.}
Data selection has been viewed as a coreset selection problem~\citep{phillips2017coresets}, which aims to find a subset of training examples that induces performance similar to training on the full dataset~\citep{toneva2018empirical, sener2018active, coleman2019selection, kaushal2019learning, xia-etal-2020-predicting, mirzasoleiman2020coresets, feldman2020neural, killamsetty2021grad, killamsetty2021retrieve, paul2021deep, meding2021trivial, mindermann2022prioritized, sorscher2022beyond,xia2023moderate}. 
These works focus on in-domain coreset selection, whereas our work involves transfer learning. 
Several prior works use pre-defined notions of useful data~\citep{gururangan2020don,chen2023skillit} or n-gram features~\citep{xie2023data} to select pre-training examples, but \ours{} more closely resembles \citet{mirzasoleiman2020coresets,wang2020optimizing, yu2020gradient, killamsetty2021grad} in its reliance on gradient information.
Our work is closest to \citet{han2023understanding}, which used model gradients to select pre-training data to improve in-context learning performance.
However, our offline approach enables extremely efficient adaptation to new downstream tasks.

\paragraph{Data attribution and influence functions.} Our formulation of influence has been used in identifying mislabeled examples~\citep{pruthi2020estimating}, analyzing memorization effects~\citep{feldman2020neural}, and deriving various interpretability insights~\citep{madsen2022post}, but it does not permit counterfactual reasoning. 
On the other hand, influence functions~\citep{hampel1974influence} can counterfactually reason about model behaviors and training data~\citep{koh2017understanding}, but some studies have found limitations~\citep{bae2022if} in its robustness~\citep{basu2020influence} and effectiveness~\citep{ilyas2022datamodels}. 
In the LLM setting, influence functions are computationally expensive~\citep{grosse2023studying}, though concurrent work in~\citet{engstrom2024dsdm} proposed using an efficient influence function estimation~\citep{park2023trak} to select pre-training data (see~\Cref{app:datamodels} for a detailed comparison of influence functions and our influence formulation). 
Our findings suggest that first-order influence approximations~\citep{pruthi2020estimating} are effective for data selection in transfer learning settings~\citep{han2023understanding}.  

\section{Discussion and Future Work}
\label{sec:discussion}
We propose a data selection algorithm \ours{} based on optimizer-aware influences (\textsection\ref{sec:prelims} and \textsection\ref{sec:influence_selection}).
\ours{} creates a datastore of effective and reusable low-dimensional gradient features to enable efficient data selection (\textsection\ref{sec:efficient_influences}). 
Experiments in \textsection\ref{sec:experiments} demonstrate the effectiveness of \ours{} compared to relevant baselines and also highlight the ability to use smaller models to select data to train larger models (\Cref{tab:main_result}).
Analysis and ablations in \textsection\ref{sec:analysis} demonstrate that our method selects more interpretable data but can be computationally expensive.
We discuss the limitations in detail in~\Cref{sec:limitations}. 

In the future, \ours{} could be effectively used for test-time adaptation, allowing the model to adapt on data selected based on the context and demonstrations provided by the users. Additionally, we could modify the data selection procedure to minimize \emph{any} differentiable metric (e.g., a surrogate for toxicity and harmfulness) instead of just cross-entropy loss.
Moreover, our experiments and concurrent findings in~\citet{engstrom2024dsdm} may prompt further investigation of the utility of gradient features as opposed to surface-form cues in data selection. 
Insights into the optimization characteristics and sequence-level gradient phenomena (\textsection\ref{subsec:inst_grad}) are likely of theoretical interest.

Our observation that tuning on more data can degrade performance on a particular task may also be intimately connected to the catastrophic forgetting behaviors observed in LLMs~\citep{lin2023speciality}.

\section*{Limitations}\label{sec:limitations}
We discuss the limitations of our approach here. 
\begin{enumerate}
    \item \textbf{Warmup training phase is required.} The warmup training phase is crucial for obtaining useful gradient features for data selection.  It is evident that we must fine-tune the base model to align with the specific data distribution we intend to select from, because directly utilizing gradients from pre-trained models like \llamasmall{} or \llamasmallchat{} is not as effective (\autoref{tab:warmup_data}). This additional step introduces increased complexity and computational load to \ours{}.
    \item \textbf{Average gradients over completion tokens.} 
The average gradient of all the completion tokens is used as a surrogate to represent the entire sequence.
In scenarios involving longer sequences in training or validation data, such as open-ended generation, this aggregation could become more ambiguous and less effective. 
In \Cref{subsec:inst_grad}, we highlight how this operation can cause anomalies in the data selection pipeline.
    \item \textbf{Minimizing loss does not monotonically increase performance.}
	Unlike standard vision settings, minimizing the validation loss (i.e., cross entropy loss) does not always improve the task performance in instruction tuning.
	This is because instruction tuning tasks can generally require the model to produce long-form generations. Although perplexity aligns well with task performance in base LLMs~\cite{xia-etal-2023-training}, but the relationship is much more obscure in fine-tuning tuning or instruction tuning~\cite{tay2021scale, zhou2023lima}.
	\ours{} is motivated by selecting datapoints that minimize the validation loss (\Cref{sec:prelims}) and this strategy often induces models with high accuracy (\Cref{tab:main_result}), but minimizing loss does not monotonically improve accuracy (see our discussion in~\Cref{sec:loss_vs_acc}).
    \item \textbf{Linearity in data selection.} The first-order approximation in \Cref{sec:prelims} ignores the effect of adding multiple datapoints together. In particular, two duplicate points would be scored equally highly and thought to doubly improve the model, though this is likely not the case. Preliminary theory in~\citet{saunshi2023understanding} discusses when influences may add linearly. However, several more sophisticated algorithms have been proposed to more flexibly model interactions across datapoints at a significantly higher computational cost~\citep{sogaard2021revisiting,guu2023simfluence}. 
    Unfortunately, such algorithms are too expensive in the language model setting.
\end{enumerate}
\section*{Acknowledgments}
We thank Logan Engstrom, Xiaochuang Han, Kaifeng Lyu, Abhishek Panigrahi, Yizhong Wang, Zhaofeng Wu, Dingli Yu, and Haoyu Zhao for their insightful discussions on the project. 
We thank Tianyu Gao, Jiayi Geng, Zirui Wang, Howard Yen, and other members of Princeton NLP for proofreading this work. 
MX is supported by the Bloomberg Data Science Ph.D. Fellowship. 
SM and SA are supported by NSF and ONR.
This research is also funded by the National Science Foundation (IIS-2211779).

\section*{Impact Statement}
This paper presents work whose goal is to advance the field of Machine Learning. There are many potential societal consequences of our work, none which we feel must be specifically highlighted here.


\bibliography{custom}
\bibliographystyle{icml2024}

\newpage
\appendix
\onecolumn

\section{Training}
\subsection{Training Datasets}
\label{app:train_data}
For training, we utilize the same four processed datasets as referenced in \citep{wang2023far}. Each of these datasets is either annotated or authored by humans. Further details about these datasets can be found in \Cref{tab:train_data}. The \flan{} and \cott{} datasets are derived from existing NLP datasets, while \dolly{} and \oasst{} consist of open-ended generation examples featuring answers written by humans. These datasets vary significantly in format, sequence length, and underlying tasks, reflecting the heterogeneity of instruction-tuning datasets. We adopt the following 'Tulu' format from the study by~\citet{wang2023far} to standardize the format of these datasets.

\begin{tcolorbox}
\textbf{\textless\textbar user\textbar\textgreater} \\
Alice's parents have three daughters: Amy, Jessy, and what's the name of the third daughter?
\\\\
\textbf{\textless\textbar assistant\textbar\textgreater} \\
The name of the third daughter is Alice
\end{tcolorbox}

\begin{table*}[h]
    \caption{Datails of training dataset from \citet{wang2023far}. Len. is short 
    for token length. }
    \centering
    \resizebox{0.96\textwidth}{!}{%
    \begin{tabular}{lrlccc}
        \toprule
        
    \textbf{Dataset}                & \textbf{\# Instance} & \textbf{Sourced from}                              & \textbf{\# Rounds} & \textbf{Prompt Len.} & \textbf{Completion Len.} \\ \cmidrule(r){1-6}
    \flan & 100,000                         & NLP datasets and human-written instructions & 1                             & 355.7                          & 31.2                               \\
    \cott   & 100,000  & NLP datasets and human-written CoTs         & 1  & 266   & 53.2   \\
    \dolly  & 15,011                          & Human-written from scratch                & 1                             & 118.1                          & 91.3                               \\ 
    \oasst & 55,668                          & Human-written from scratch                & 1.6                           & 34.8                           & 212.5     \\ \bottomrule                        
    \end{tabular}}
    \label{tab:train_data}
\end{table*}

\subsection{Training Details}
\label{app:training_setup}
All experiments were conducted using the parameter-efficient fine-tuning method LoRA~\citep{hu2021lora}. We employed a learning rate scheduler with linear warm-up and cosine decay, reaching a peak learning rate of $2\times10^{-5}$. A batch size of 128 was used, and training was carried out for 4 epochs across all selected datasets. We initially conducted experiments to align the number of training steps between small and large datasets. However, increasing the number of epochs on small datasets did not yield performance improvements. Consequently, we opted for a fixed number of epochs in all experiments.

For the LoRA module, we specified a rank of 128, an $\alpha$ value of 512, a dropout rate of 0.1, and learned LoRA matrices for all attention matrices. This configuration resulted in 135 million trainable parameters for \llamasmall{}, constituting $1.95\%$ of the original \llamasmall{} model. For \mistral{}, there were 109 million trainable parameters, accounting for $1.48\%$ of the original \mistral{} model. Lastly, the \llamalarge{} model comprised 209 million trainable parameters, representing $1.59\%$ of the original \llamalarge{} model.

In each experiment, we performed three trials using distinct random seeds. When employing random selection approaches, this entailed selecting three different random subsets from the training dataset. In the case of our approach (\ours{}), it involved conducting warmup training with various subsets of the training data and subsequently selecting different subsets for each trial from each warmup-trained model. We used consistent optimization seeds across all experiments.

\section{Evaluation Details}
\label{app:eval}
We follow \citet{wang2023far} to evaluate the performance of the models on the target tasks. For \mmlu{}, we measure the 5-shot accuracy of the test set averaged across 57 subtasks. For \tydiqa{}, we measure the 1-shot macro-averaged F1 score across all 11 languages. We adopt the gold-passage setup where one passage containing the reference answer is provided to the model. For \bbh{}, we report the average 3-shot exact match score across all tasks. Chain-of-thought reasoning is provided in each in-context learning example to prompt the model to generate chain-of-thought reasoning traces for test examples. We evaluate on the validation set $\dval$ (the same reference set used for data selection) at the end of each epoch and select the best checkpoint to evaluate on the final test set for each experiment. Note that this procedure might introduce some bias to the final test set, given that the validation set is relatively small (e.g., \tydiqa{} only has 9 validation examples in total). However, in practice, we don't find this bias to affect the comparisons between different methods.

\section{Baseline Details}
For baseline approaches that involve stochasticity (i.e., all except BM25), we perform three runs with different random seeds and report the average performance and standard deviation. 

\label{app:baselines}
\subsection{BM25}
We implement the BM25 baseline with the \texttt{rank\_bm25}\footnote{\url{https://github.com/dorianbrown/rank\_bm25}} package. We employ the same approach to aggregating scores across subtasks as~\Cref{eq:matching}, except using the BM25 scores instead of the gradient similarity scores. 

\subsection{DSIR}
\label{app:dsir}
DSIR~\cite{xie2023data} offers a lightweight approach to select data  based on specific target distributions using n-gram features. However, we find it unsuitable for our specific setting. The primary issue stems from the difficulty in learning a reliable n-gram model with the limited data available in our target distribution, leading to a sparse n-gram table. We hypothesize that this sparsity arises from the trade-off between bucket size and estimation accuracy for the source and target distribution: larger buckets result in a highly sparse table with unreliable estimations for data selection, while smaller buckets lead to more hash collisions for the data in the source distribution. In our empirical observations, we noticed that DSIR tends to prefer shorter sequences, which can harm performance (see~\Cref{tab:normalized_grads} for similar evidence).
The tendency of DSIR to select short sequences is also mentioned in Appendix J of~\citet{xie2023data}, where the authors impose heuristic quality filters on top of the DSIR-selected datasets.

\section{More Experiment Results}
\subsection{Comparing SGD and Adam Optimizers for Instruction Tuning}
\label{app:adam_sgd_ft}
The Adam optimizer~\citep{kingma2015adam} is commonly employed for fine-tuning large language models. 
However, SGD is a more desirable optimizer due to its lower memory usage; moreover, the influence formulation we introduce in \textsection\ref{sec:prelims} applies only to SGD.
Here, we verify that using SGD to instruction tune models results in markedly worse performance~\Cref{tab:adam_sgd_ft}.
Our findings consistently demonstrate that using the Adam optimizer yields better results across all three datasets. This validates the widespread practice of utilizing the Adam optimizer for fine-tuning large language models.

\begin{table}[h]
\caption{Comparing using the SGD and Adam optimizers for instruction tuning. We use a random subset of $25\%$ data for comparison. We find that using the Adam optimizer consistently outperforms SGD on all three datasets.}
    \centering
\begin{tabular}{lcccc}
    \toprule
          & \textbf{\mmlu}           & \textbf{\tydiqa} &  \textbf{\bbh} & \textbf{Average} \\ \midrule
    \textbf{SGD}   & 45.5 \std{0.3} & 47.3 \std{1.2} & 39.2 \std{0.5} & 44.0 \\
    \textbf{Adam}  & 47.2 \std{0.8} & 53.1 \std{1.3} & 40.4 \std{1.5} & \textbf{46.9} \\ \bottomrule
\end{tabular}

\label{tab:adam_sgd_ft}
\end{table}

\subsection{Selecting Data for SGD, SignGD, and Adam Gradients}
\label{app:adam_sgd}

In order to compute the influence when training with the Adam optimizer (\Cref{def:influence}), we need to access the Adam optimizer update $\Gamma$.
$\Gamma$ relies on the moving moment estimates that Adam tracks.
In the early stages of training, \citet{malladi2022sdes} proved that the parameter updates induced by Adam can be approximated well by the sign of the gradient (this optimizer is known as SignGD).
Intuitively, this is because the moment estimates $\vm$ and $\vv$ are close to their initialization and thus essentially perform coordinate-wise normalization on the gradient.
Later empirical and theoretical in \citet{malladi2023kernel} suggested that fine-tuning language models occurs in the early stages of Adam, so signGD may be a decent approximation of fine-tuning LLMs with the Adam optimizer.

In the current paper, we obtain $\Gamma$ through the optimizer states computed over the course of a short warmup training period.
Here, we ablate our stated approach against approximating $\Gamma$ as $\textrm{sign}(\grad{\cdot}{\vtheta})$.
One may also think this SignGD formulation could circumvent the gradient norm problems reported in \textsection\ref{subsec:inst_grad}.

One of our key contributions is to extend the influence formulation in \textsection\ref{sec:prelims} to operate with the Adam optimizer.
However, due to the approximations described above, the influences we compute in \Cref{def:influence} may not be as accurate as we hope.
In that case, we may be better off using the more precise influences provided by the SGD formulation in~\Cref{eq:infsgd}.

We empirically compare the efficacy of these three influence formulations (SGD, SignGD, and Adam) for data selection and present the results in Table \ref{tab:adam_sgd}. 
In this setup, we perform warmup training using all the available training data ($\mathcal{D}_{\mathrm{warmup}} = \mathcal{D}$) and select the top 5\% of the data for instruction tuning. 
In summary, we find that using Adam influence formulation generally matches or outperforms the SGD and SignGD formulations, demonstrating the utility of our optimizer-aware influence formulation in~\Cref{def:influence}.


\begin{table*}[h]
        \caption{Comparing the data selection capabilities of the SGD, SignGD, and Adam~(\Cref{def:influence}) formulations of influence. We find that using the Adam formulation generally matches or outperforms using the SGD and SignGD ones, showing the utility of our optimizer-aware influence formulation.}
    \centering
    \resizebox{0.4\textwidth}{!}{
    \begin{tabular}{lccccc}
    \toprule
     & \textbf{Random} & \textbf{SGD} & \textbf{SignGD} & \textbf{Adam} \\ \midrule
    \textbf{\mmlu} & 46.5 & 50.9 & 49.7 & \textbf{51.6} \\
    \textbf{\tydiqa} & 52.7 & \textbf{58.1} & 54.8 & 57.9 \\
    \textbf{\bbh} & 38.9 & 40.0 & 39.0 & \textbf{41.9} \\ \midrule
    \textbf{Avg.} & 46.0 & 49.7 & 47.8 & \textbf{50.5} \\
    \bottomrule
    \end{tabular}}

    \label{tab:adam_sgd}
\end{table*}


\subsection{Comparison to Base Models, and Training with Validation Data}
\label{app:valid}

\begin{table}[h]
    \centering
        \caption{Evaluation results with the base model (without instruction tuning), and with instruction tuning on the validation set. Finetuining on the validation set alone rarely improves the performance on the test set.}
    \resizebox{\textwidth}{!}{
    \begin{tabular}{@{}lcccc|cccc|cccc@{}}
    \toprule
    & \multicolumn{4}{c|}{\textbf{\llamasmall}} & \multicolumn{4}{c|}{\textbf{\llamalarge}} & \multicolumn{4}{c}{\textbf{\mistral}} \\
    \cmidrule(lr){2-5} \cmidrule(lr){6-9} \cmidrule(lr){10-13}
    & \textbf{Base} & \textbf{Valid.} & \textbf{Rand. 5\%} & \textbf{\ours{} 5\%} & \textbf{Base} & \textbf{Valid.} & \textbf{Rand. 5\%} & \textbf{\ours{} 5\%} & \textbf{Base} & \textbf{Valid.} & \textbf{Rand 5\%} & \textbf{\ours{} 5\%} \\ 
    \cmidrule(lr){2-5} \cmidrule(lr){6-9} \cmidrule(lr){10-13}
    \mmlu & 45.6 & 47.3 \std{0.3} & 46.5 \std{0.5} & 50.2 \std{0.5} & 55.3 & 54.8 \std{0.4} & 53.4 \std{0.1} & 54.0 \std{0.7} & 62.4 & 58.8 \std{0.4} & 60.0 \std{0.1} & 61.8 \std{0.4} \\
    \tydiqa & 46.4 & 52.8 \std{2.5} & 52.7 \std{0.4} & 56.2 \std{0.7} & 53.1 & 50.8 \std{1.4} & 53.0 \std{1.3} & 54.6 \std{0.3} & 49.6 & 45.4 \std{4.3} & 56.9 \std{0.2} & 60.3 \std{2.4} \\
    \bbh & 38.3 & 36.5 \std{0.9} & 38.9 \std{0.5} & 41.5 \std{0.6} & 46.2 & 45.8 \std{0.9} & 47.0 \std{1.6} & 50.6 \std{0.6} & 54.9 & 55.5 \std{0.7} & 54.5 \std{0.1} & 56.8 \std{1.0} \\
    \bottomrule
    \end{tabular}}
    \label{tab:icl_result}
    \end{table}

\paragraph{Instruction tuning could degrade performance.} In \Cref{tab:icl_result}, we compare results from instruction tuning to the results directly from the base model without instruction tuning. Athough instruction tuning generally enhances performance, there are instances where it may result in a decline. A notable example is observed in the \mmlu{} dataset, where instruction tuning using the 5\% data subset chosen by \ours{} yields slightly inferior results compared to the base model. This suggests that the chosen data may not significantly improve the performance of the target task for some models, especially more capable ones such as \llamalarge{} and \mistral{}. Additionally, it could imply that the pool of training data we're selecting from lacks further beneficial examples for the specified task. Experimenting with a larger pool of instruction tuning could possibly benefit the target task more. Despite this, it is important to note that \ours{} remains competitive to other approaches. By scaling up the compute to have a better estimate of the influence, we expect that the performance of \ours{} will further improve, as indicated in~\Cref{subsec:efficiency}.

\paragraph{Training with validation data rarely helps.} An additional straightforward baseline involves using the few-shot validation data for training. Specifically, we train the model for 100 steps with the standard hyperparameters, and evaluate the model on the validation set for every 10 steps, and report the test score of the best checkpoint. The results of this approach are shown in \autoref{tab:icl_result}. Here, we compare it with the performance of the standard \llamasmall{} model and the outcomes when fine-tuning the model with either $5\%$ of data chosen at random or $5\%$ selected by \ours{}. We observe that training solely with in-domain validation data rarely provides a benefit over not training the model at all, and it even surpasses the performance achieved by randomly selecting $5\% (13,000)$ data points. Nevertheless, \ours{} demonstrates a superior ability to identify data points more relevant to the target task, significantly outperforming the strategy of training with only validation data. 

\subsection{Detailed Results on Using Different Projection Dimensions}
\label{app:dimension} 
Recall that the Johnson-Lindenstrauss lemma~\citep{Johnson1984ExtensionsOL} states that random projections preserve inner products (e.g., the influence formulation in~\Cref{def:influence}) as long as the projected dimension $d$ is large enough. Here, we ablate against different projection dimensions to find what empirical choice of $d$ is ``large enough''. In this setup, we perform warmup training using all the available training data ($\mathcal{D}_{\mathrm{warmup}} = \mathcal{D}$) and select the top 5\% of the data for instruction tuning. We present the detailed results of projecting the LoRA gradients to different dimensions including $1024$, $2048$, $4096$ and $8192$ in \Cref{tab:dim}. We find that even using a small dimension still outperforms the random selection performance. Using a larger dimension for projected gradients is better. This signifies that there is still potential to improve the performance by using a larger projection dimension, but one must trade off this consideration with the increased computational and memory cost. Note that due to computational constraint, we only run experiment once with one fixed seed.

    \begin{table}[h]
        \caption{Performance of using different projected gradient dimensions for data selection. We find that even using a small dimension still outperforms the random selection performance. Using a larger dimension for projected gradients is better.}
        \centering
        \resizebox{0.45\columnwidth}{!}{%
            \begin{tabular}{lccccc}
            \toprule
                             & \multicolumn{1}{l}{\textbf{}} & \multicolumn{4}{c}{\textbf{Projected Gradient Dimension}}     \\ \cmidrule(lr){3-6}
            \textbf{}        & \textbf{Random}               & \textbf{1024} & \textbf{2048} & \textbf{4096} & \textbf{8192} \\ 
            \cmidrule(lr){1-1} \cmidrule(lr){2-2} \cmidrule(lr){3-6}
            \textbf{MMLU}    & 46.5                          & 50.7          & \textbf{51.2} & 50.5          & 51.1          \\
            \textbf{TydiQA}  & 52.7                          & 55.3          & 56.3          & \textbf{56.8} & 56.6          \\
            \textbf{BBH}     & 38.9                          & 39.3          & 39.0          & 40.4          & \textbf{41.3} \\ \midrule
            \textbf{Average} & 45.2                          & 48.4          & 48.8          & 49.2          & \textbf{49.7} \\ \bottomrule
            \end{tabular}
        }
        \label{tab:dim}
    \end{table}

\subsection{Results on Pythia Models}
\label{subsec:pythia}
To explore how the data selection model in \ours{} scales with increasing model capabilities, we perform experiments using LESS with a series of Pythia models~\citep{biderman2023pythia} of sizes 14M, 410M, and 1B to select data for training models of 14M, 410M, 1B, 6.9B, and 12B. Due to constraints on time and computational resources, we conduct these experiments to select data for the TydiQA benchmark using only one seed and present the results in~\Cref{tab:ft_comparison}.

\begin{table}[h]
    \centering
    \resizebox{0.8\columnwidth}{!}{%
    \begin{tabular}{lcccccc}
    \toprule
    & & & \multicolumn{3}{c}{\textbf{Data selection with LESS by}} & \\
      \cmidrule{4-6} 
    & \textbf{base (w/o FT)} & \textbf{Random} & \textbf{Pythia-14m} & \textbf{Pythia-410m} & \textbf{Pythia-1b}  & \textbf{Full} \\
    \midrule
    \textbf{Amount of FT data} & 0\% & 5\% & 5\% & 5\% & 5\% & 100\% \\ \midrule \midrule
    \textbf{Pythia-14m} & 0.0 & 0.0 & 0.0 & 0.1 & 0.0 & 0.0 \\
    \textbf{Pythia-410m} & 3.8 & 7.8 & 8.7 & 9.4 & 8.8 & 8.6 \\
    \textbf{Pythia-1b} & 9.2 & 12.3 & 12.9 & 11.5 & 17.2 & 13.7 \\
    \textbf{Pythia-6.9b} & 17.1 & 27.2 & 27.2 & 28.0 & 30.9 & 35.5 \\
    \textbf{Pythia-12b} & 17.4 & 14.9 & 17.7 & 19.4 & 32.6 & 33.2 \\ \midrule
    \textbf{Average} & 9.5 & 12.4 & 13.3 & 13.7 & 17.9 & 18.2 \\
    \midrule \midrule
    \textbf{\llamasmall} & 46.4 & 52.7 & 48.7 & 53.0 & 51.7 & 54.0 \\
    \bottomrule
    \end{tabular}}
    \caption{Performance comparison of using different Pythia models for data selection.}
    \label{tab:ft_comparison}
\end{table}

Firstly, LESS consistently performs well with Pythia models, almost always surpassing the baseline of selecting a random 5\% of the data. Furthermore, the approach of using one model to select data for training another (i.e., LESS-T) proves effective within the Pythia model family. Interestingly, small models such as the 14M can be effective data selectors despite being unable to solve TydiQA by themselves. This suggests that LESS can be made highly efficient using very small models for the gradient datastore, as they can effectively select data for training larger models. However, larger models demonstrate even greater effectiveness as data selectors, as illustrated in the last row of the table. Lastly, the data selection by Pythia models does not translate effectively to Llama models. Data selected by Pythia models fails to outperform random selection when used to train Llama2 models. Conversely, data chosen by Llama2 models effectively trains Mistral models. This indicates that the success of data transfer between models can be unpredictable, particularly when the capabilities of the models differ significantly like Pythia and Llama2.

\subsection{Results on GSM8K and TruthfulQA}
\label{subsec:gsm8k_truthfulqa}
We follow reviewers' suggestions to run experiments on datasets GSM8K~\citep{cobbe2021training} and TruthfulQA~\citep{lin2021truthfulqa} that are more considered as out of domain with regard to the training data. We evaluate GSM8K using the 8-shot setting with chain of thought, following the methodology of \citet{wang2023far}. Similarly, we use random 10 examples from each dataset as the validation set. We present the results in \Cref{tab:out_of_domain}. In summary, we find that LESS consistently outperforms random selection by a large margin. However, selecting only 5\% of the data to solve GSM8K does not outperform using the entire dataset. This is because the COT portion of our training data contains many in-domain examples for GSM8K, so it is likely not a good idea to use only a small subset of the available data to train the model for that task. LESS’ utility in selecting data for TruthfulQA indicates its potential to make models not only more capable but also more faithful and truthful.

\begin{table}[h]
    \centering
    \begin{tabular}{@{}lcc@{}}
    \toprule
    \textbf{}                     & \textbf{GSM8K-COT} & \textbf{TruthfulQA} \\ 
    & (Exact Match) & (MC2) \\ \midrule
    \textbf{w/o Finetuning (0\%)} & 14.0           & 39.2                \\
    \textbf{Random (5\%)}         & 17.0           & 41.0                \\
    \textbf{LESS (5\%)}           & 21.0           & 48.0                \\
    \textbf{Full (100\%)}         & 30.5           & 44.0                \\ \bottomrule
    \end{tabular}
    \label{tab:out_of_domain}
\end{table}

\section{Nuances in Algorithm Design}\label{sec:nuance}
In~\Cref{sec:prelims,sec:influence_selection}, we outline how first-order approximations of influence functions can be used to design optimizer-aware data selection algorithms.
Here, we discuss some of the nuances in designing such an algorithm for the practical hyperparameter settings used in instruction tuning.

\paragraph{Batch Size.}
The per-step influence and trajectory-wise influence of a training datapoint $\vz$ on a validation datapoint $\vz'$ is characterized under the assumption that the batch size is $1$.
Standard training setups use a larger batch size, but it is difficult to characterize how training on a batch of examples can change $\ell(\vz';\vtheta)$.
However, under the assumption that the first-order approximation is sufficiently powerful to describe training dynamics, we can directly use the linearity of the gradient to derive the influence of a single datapoint within a batch.
In particular, when training with SGD on a batch $B_t$ of size $B$, we can write:
\begin{align*}
	\ell(\vz';\vtheta^{t+1}) - &\ell(\vz';\vtheta^t) \approx -\eta_t\langle\grad{\vz'}{\vtheta^t}, \grad{B_t}{\vtheta^t}\rangle \\
	&= -\frac{\eta_t}{B} \sum_{\vx\in B_t} \langle \grad{\vz'}{\vtheta^t}, \grad{\vx}{\vtheta^t}\rangle
\end{align*}
An analogous formula follows straightforwardly for Adam.
The $1/B$ factor is shared across all datapoints and thus does not affect the ranking of the matching scores.
Therefore, selecting examples as if the batch size were $1$ is the same as selecting examples for a larger batch size.
Intuitively, this is because the interaction between examples within a batch must be negligible in order for the first-order approximation to be accurate.
We remark that the equal division of the influence across constituents of a mini-batch cannot circumvent the gradient norm problem presented in \textsection\ref{subsec:inst_grad}. 
We discuss this point in more detail in \Cref{sub:grad_norm}.

\paragraph{Learning Rate Schedule.}
Instruction tuning generally uses a linearly decaying learning rate schedule, which means that each example (or batch) within an epoch has a different learning rate depending on the time step that it was sampled.
The formulation in~\Cref{def:influence} may then select a datapoint solely because it would be sampled earlier in the training run (i.e., with a larger $\eta_t$).
We want our method to yield an interpretable dataset, so we instead use the average epoch-wise learning rate as $\eta_t$. 

\section{Comparison to ``Kernel Behavior''~\citep{malladi2023kernel, jacot2018neural}.}\label{sec:kernel}
The design of \ours{} relies on the assumption that the first-order approximation of loss dynamics is sufficiently accurate (see~\Cref{sec:prelims}).
A related hypothesis is the idea that the instruction tuning optimization procedure exhibits ``kernel behavior''.\footnote{This was called the ``kernel regime'' in~\citet{jacot2018neural} when training the model from a random initialization.}
Here, we highlight the nuanced difference between kernel behavior and the current assumption, emphasizing why kernel behavior is difficult to measure in the instruction tuning setting.

\paragraph{Formal Difference} 
Kernel behavior describes the gradient of the output of the model (i.e., the logits) with respect to the model parameters. 
In contrast, the first-order approximation in~\Cref{sec:prelims} is for the gradient of the \emph{loss} with respect to the parameters. 
These two quantities are related by the derivative of the loss with respect to the output of the model.
Thus, conceptually, kernel behavior and the first-order approximation of the loss dynamics are similar, but the two phenomena correspond to different formal descriptions of model training.
Theory in~\citet{yang2021tensor,malladi2023kernel} indicates that there is some subtlety in whether one phenomenon implies the other holds.
In particular, most kernel-based analyses minimize the mean squared error loss instead of the cross-entropy loss, the latter of which introduces technical complications in the analysis but is the standard choice for instruction tuning. 

\subsection{Does instruction tuning exhibit kernel behavior?}
\citet{malladi2023kernel} suggested that fine-tuning on GLUE-style classification tasks can sometimes exhibit kernel behavior because the pre-trained model is already fairly strong at the  downstream task.
This assumption crucially relied on appending a very good prompt to the downstream task input, so that labeling the input amounted to predicting the next word.  
With such a prompt, fine-tuning on the downstream task amounted to the same as continued pre-training.
One could argue that the instruction in instruction tuning provides a prompt for the completion, but the completion can be much longer than the single-word labels used in classification tasks. 

\begin{figure}[t]
    \includegraphics[width=\columnwidth]{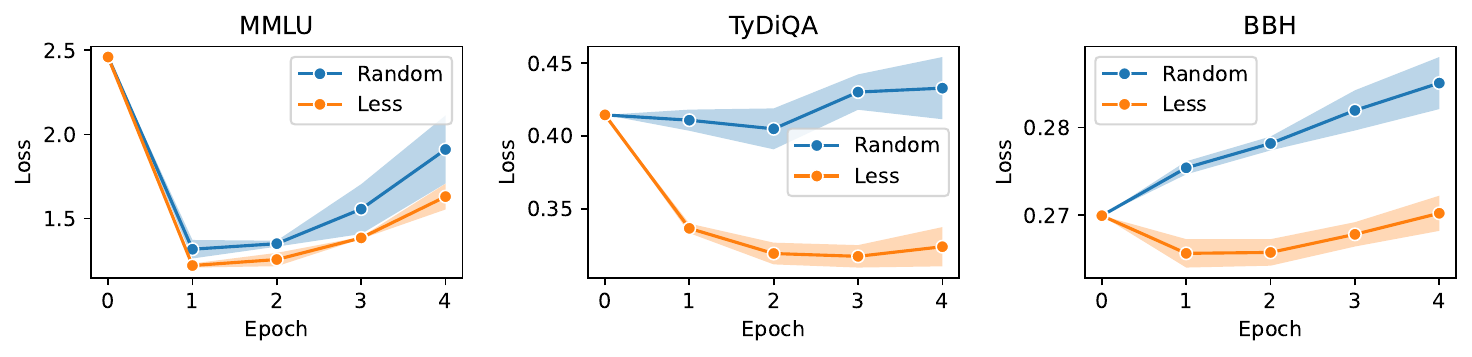}
    \caption{Validation loss of random selection and \ours{} across checkpoints for \llamasmall{}.}
    \label{fig:llama_loss}
\end{figure}

\paragraph{Gradient features change substantially at the start of instruction tuning.}
We perform ablations comparing gradient features collected from the pre-trained model and features collected over the course of the warmup training period (\Cref{tab:number_ckpts}). 
We find that it is important to incorporate gradients taken over the course of the warmup training, which suggests that the gradient features change substantially at the start of instruction tuning.
On the other hand, results in~\Cref{tab:warmup_data} demonstrate that lengthening the warmup period yields marginal gains in performance, suggesting that the gradient features don't change much in the intermediate and late phases of instruction tuning.
We emphasize that especially in the case of the cross entropy objective, it is possible for the gradient features in \Cref{def:influence} to change substantially even if the model exhibits kernel behavior.
It is thus difficult to relate the loss curves in~\Cref{fig:llama_loss} to the evolution of the empirical neural tangent kernel over the course of instruction tuning.

\paragraph{Direct verification is too challenging.}
Kernel behavior is called as such because the dynamics are described by a fixed $|\dfull| \times |\dfull|$ kernel, where each entry contains the inner product of the gradients of the model output with respect to the parameters for different datapoints.
Subtly, this matrix must be constructed for every logit, so the computational cost scales with the number of classes in the dataset. 
Past works that instantiate this kernel have focused on vision tasks with relatively few classes~\citep{arora2019exact,yu2020enhanced,wei2022more} or prompt-based text classification~\citep{malladi2023kernel}, which only considers a label word for each class in simple downstream tasks.
On the other hand, instruction tuning requires a $|\gV|$-dimensional vector of logits from the model, where $\gV$ is the vocabulary containing, for example, roughly $32,000$ tokens.
Additionally, each partial context must be treated as its own input, making $|\dfull|$ the length of each sequence multiplied by the number of sequences in the dataset.
Therefore, computing the kernel for instruction tuning is computationally infeasible, given the massive size of $\dfull$ and the large number of output classes.

\section{Comparison to Data Models~\citep{ilyas2022datamodels}}\label{app:datamodels}
Concurrent work in~\citet{engstrom2024dsdm} applies the datamodels framework~\citep{ilyas2022datamodels} to select pre-training data to induce strong performance on target downstream tasks.
Fitting a datamodel generally requires repeated training of a model to fit a mapping between the selected data and the resulting accuracy of the trained model, but~\citet{park2023trak} reframed the problem  to simulate this repeated training procedure at a much lower computational cost.
Datamodels and its related techniques are inspired by influence functions~\citep{koh2017understanding}, which permit counterfactual reasoning over training (e.g., ``what would have happened if I omitted this training datapoint?'').
On the other hand, the influence formulation in~\citet{pruthi2020estimating} can only counterfactually reason over one optimization step at a time, not the entire training procedure.

In this section, we describe the how the data selection method in~\citet{engstrom2024dsdm}, which was applied to the pre-training setting, relates to the considerations and observations reported in this paper.
\begin{enumerate}
	\item \textbf{Reframing data selection as optimization}: Both works reframe the data selection problem as a straightforward optimization problem, whereby one chooses data to maximally improve a given metric. This approach is in contrast to prior works that rely on pre-defined notions of useful data~\citep{gururangan2020don,chen2023skillit} or surface-form features~\citep{xie2023data}. Instead, both works can be considered efficient approximations of meta-optimization approaches~\citep{nguyen2021dataset}.
	\item \textbf{Approximating training dynamics}: Given the considerable computational cost of training LLMs, both works have to approximate the training dynamics of the model in order to efficiently estimate the effect of including a datapoint. \ours{} relies on the assumption that the first-order approximation of the loss dynamics is accurate, whereas \citet{engstrom2024dsdm} relies on a first-order approximation of how the model output changes during training. The latter is closer to the kernel view described in the previous section, though we note that the output used in datamodels generally involves the margin, which is not well-covered by kernel theory. An implicit assumption of both of these techniques is that the influences of selected datapoints add linearly (i.e., selecting two equally highly scored datapoints will doubly improve the model performance). Theory in~\citet{saunshi2023understanding} provides a start at understanding when this linearity holds, but it may not apply to all settings. For example, in the case where the same datapoint is duplicated in the training set, we don't expect a marked improvement from the inclusion of the duplicate. 
	\item \textbf{Proxy models and selection models}: Both works require performing initial training in a smaller, computationally efficient setup before performing data selection. \citet{engstrom2024dsdm} trained smaller so-called proxy models to fit the datamodel, whereas \ours{} requires a short warmup training period using the selection model in order to compute the influence in~\Cref{def:influence}.
	\item \textbf{Sequence selection versus token selection}: In \textsection\ref{subsec:inst_grad}, we describe how selecting sequences instead of individual tokens requires an adjustment to the influence formulation. The token versus sequence consideration also appears in~\citet{engstrom2024dsdm}, where the method has to average the model output over the entire sequence before using it in the data model. Because~\citet{engstrom2024dsdm} considers the pre-training setting, where all input sequences are of the same (maximal) length, we cannot be sure if their proposed methodology circumvents the sequence length bias described in~\textsection\ref{subsec:inst_grad}.   
\end{enumerate}

\section{Sequence Level Gradients in Instruction Tuning}
\label{sub:grad_norm}
\begin{figure}
    \includegraphics[width=\columnwidth]{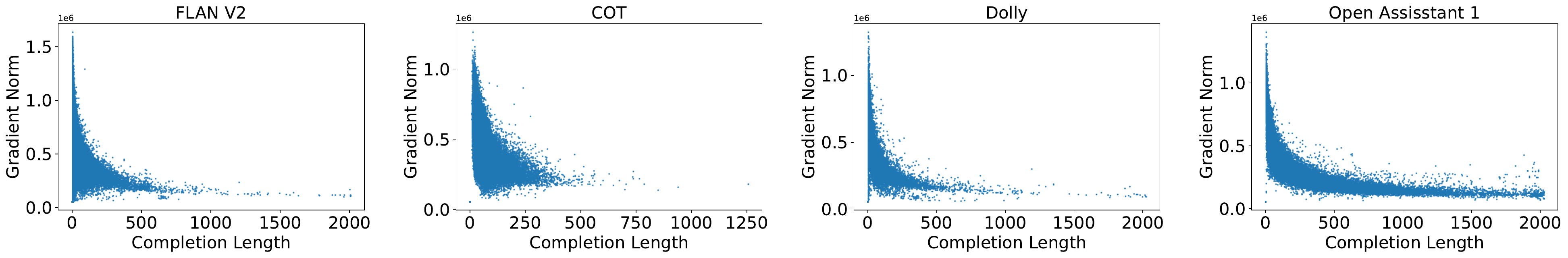}
    \caption{The norm of the update in the Adam optimizer for instruction tuning examples negatively correlates with the length of the completion.}
    \label{fig:grads_length_adam}
\end{figure}

In~\Cref{subsec:inst_grad}, we discussed that sequence level gradients in instruction tuning exhibit odd properties because they are the average of the token gradients in the completion.
In particular, \Cref{fig:grads_length} demonstrates that across datasets, the gradient norm is strongly anti-correlated with the length of the completion.
Similarly, \Cref{fig:grads_length_adam} shows that the norm of the update $\|\Gamma\|$ (see~\Cref{def:influence}) decays rapidly with length.
Therefore, using the straightforward inner product formulation in~\Cref{sec:prelims} would bias \ours{} to select short sequences for the training dataset. 
Intuitively, training on only short sequences would prevent the model from learning to perform long-form generations, as is required for many tasks.
Ablation experiments in \Cref{tab:normalized_grads} demonstrate that using inner products for \ours{} results in a dataset with much shorter sequences.
Training on this dataset induces markedly worse performance than using the cosine similarity and sometimes worse performance than random selection.
\citet{wang2020optimizing} similarly reported that using dot products for data selection in the machine translation setting was less stable than using cosine similarity.
 
 Importantly, this phenomenon is not a failure of the influence formulation in~\Cref{sec:prelims}.
Rather, it suggests that we ought to perform token-wise data selection (e.g., only averaging the loss over a few tokens within each sequence), because some gradients within a sequence are helpful and others are not.
At the time of writing, such fine-grained selection is computationally intractable for LLMs, but additional approximations and implementation accelerations may make it possible.

\subsection{Theoretical View}
\begin{figure}
    \includegraphics[width=\columnwidth]{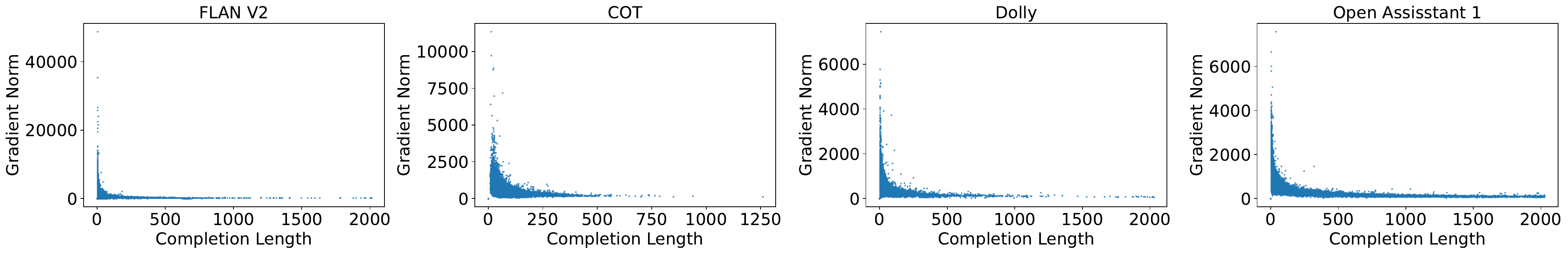}
    \caption{Gradient norms of instruction tuning examples negatively correlates with the length of the completion.}
    \label{fig:grads_length}
\end{figure}

Theoretically, the relationship between sequence-level gradients and token gradients is quite interesting.
Suppose token gradients were equal to a deterministic sequence-level gradient (e.g., determined by topic) plus some i.i.d. Gaussian noise.
This is analogous to the widely accepted theoretical model of example gradients being equal to the full batch gradient with additive noise.
Then, as the sequence length grows, we would have more independent samples of the token gradient noise, and the average token gradient should concentrate to the deterministic sequence-level gradient.  
This would yield the trends shown in~\Cref{fig:grads_length} but it is difficult to believe the gradient noise is i.i.d. because the tokens are not sampled i.i.d. from the corpus. 
Tokens within a context are semantically related to one another, and constructing a suitable noise model with appropriate correlations that permits analysis appears to be difficult. 
One possible approach is to assume a generative model for language that assign a latent topic vector to each sequence and generate tokens accordingly, as was done in~\citet{arora2016latent}.
Then, the token gradients can be modeled as correlated.

\begin{table}[h!]
    \centering
    \caption{Task performance of using normalized gradients and vanilla gradients. We find that using vanilla gradients to select data leads to worse performance than normalized gradients. We also find that the length of the selected data is much shorter when using vanilla gradients, due to the fact that shorter sequences have larger gradient norms (\Cref{fig:grads_length,fig:grads_length_adam}).}
    \begin{tabular}{@{}lcccccc@{}}
    \toprule
    & \multicolumn{2}{c}{\textbf{Random}}    & \multicolumn{2}{c}{\textbf{Vanilla Gradients}} & \multicolumn{2}{c}{\textbf{Normalized Gradients}} \\ 
           \cmidrule(lr){2-7}
           & \textbf{Performance}     & \textbf{Length}    & \textbf{Performance}     & \textbf{Length}     & \textbf{Performance}       & \textbf{Length}      \\
           \cmidrule(lr){2-3} \cmidrule(lr){4-5}  \cmidrule(lr){6-7}
    \mmlu   & 46.5  & 105.4  & 47.3                    & 11.3               & 50.2                       & 15.1                \\
    \tydiqa & 52.7 & 105.4  & 52.3                    & 19.7               & 56.2                       & 44.4                \\
    \bbh    & 38.9 & 105.4  & 37.4                    & 19.0               & 41.5                       & 101.2               \\ \bottomrule
    \end{tabular}
    \label{tab:normalized_grads}
\end{table}

\section{Loss vs Accuracy}
\label{sec:loss_vs_acc}
Our motivation in \Cref{sec:prelims} is to select datapoints that will most dramatically reduce the validation loss.
However, for many language modeling tasks, it is unclear how minimizing the cross entropy interacts with the task accuracy.
In particular, many complex tasks require the model to generate long-form sequences of text, and it is not clear how well long-form generation capability correlates with accuracy.

\begin{figure}[t]
    \includegraphics[width=\columnwidth]{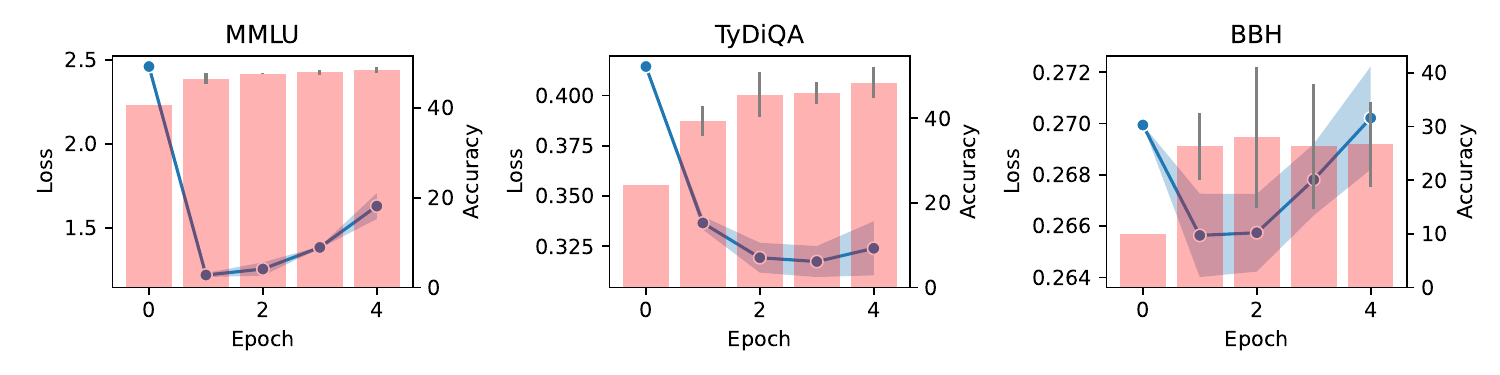}
    \caption{Validation loss (Blue curve) and accuracy (Red bar) across checkpoints for \llamasmall{}.}
   \label{fig:llama_loss_acc}
\end{figure}

\begin{figure}[t]
    \includegraphics[width=\columnwidth]{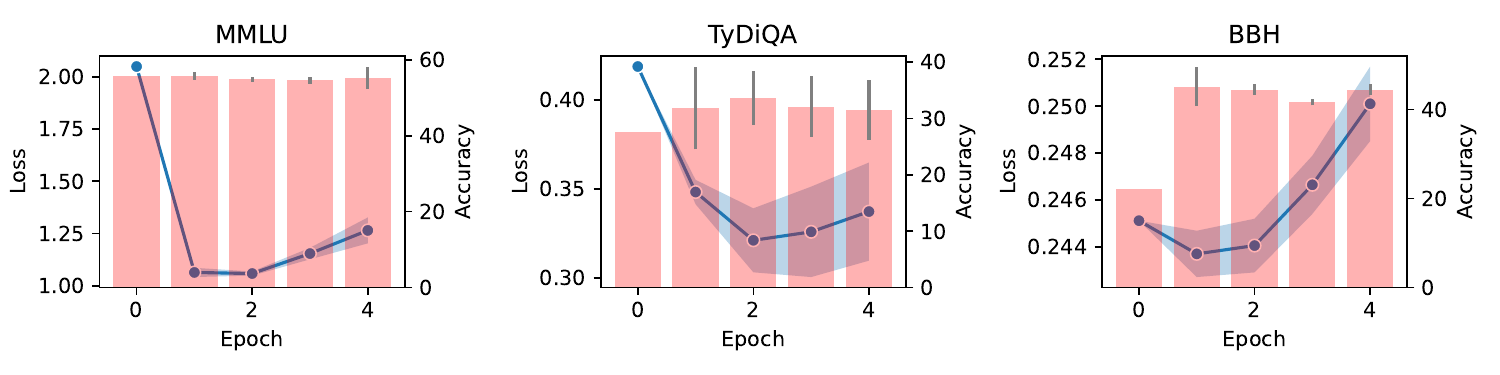}
    \caption{Validation loss (Blue curve) and accuracy (Red bar) across checkpoints for \llamalarge{}.} 
   \label{fig:llama13_loss_acc}
\end{figure}

\begin{figure}[t]
    \includegraphics[width=\columnwidth]{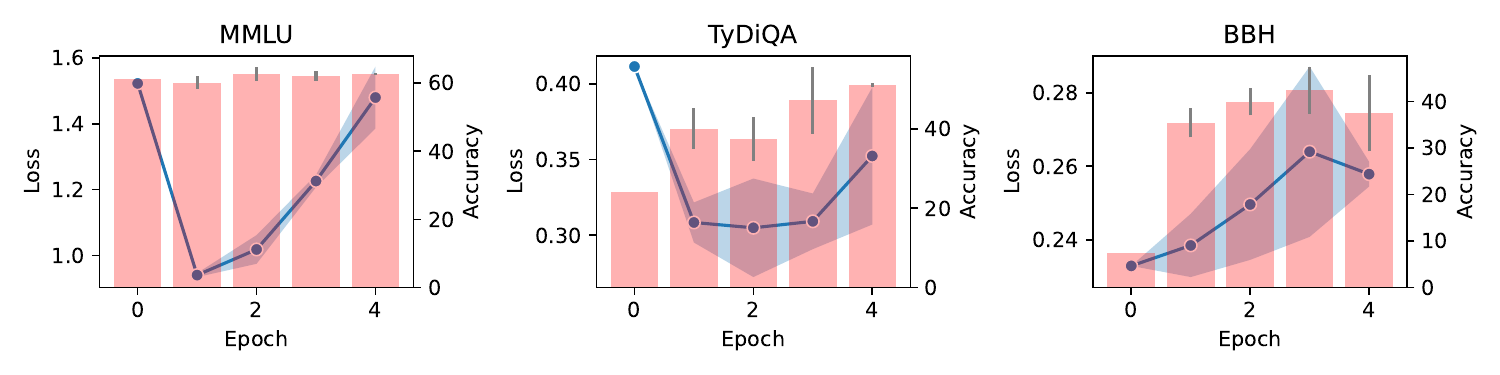}
    \caption{Validation loss (Blue curve) and accuracy (Red bar) across checkpoints for \mistral{}.}
   \label{fig:mistral_loss_acc}
\end{figure}

In~\Cref{fig:llama_loss_acc,fig:llama13_loss_acc,fig:mistral_loss_acc}, we provide evidence across model scales and tasks that the minimizing the validation loss does not always correspond to increasing the accuracy. 
A similar phenomenon was observed in~\citet{zhou2023lima,tay2021scale}.
Despite its reliance on the cross entropy loss, \ours{} boosts accuracy over relevant baselines.  
Overall, we suggest that optimizing alternative objectives may yield better accuracy, but we leave exploring such ideas to future work.
We note that any objective function can be substituted into the derivation in~\Cref{sec:prelims}, so the ideas used in developing \ours{} are easily extensible.
    
\section{Gradient Similarity Analysis}
\begin{figure*}[h]
    \centering
    \includegraphics[width=\textwidth]{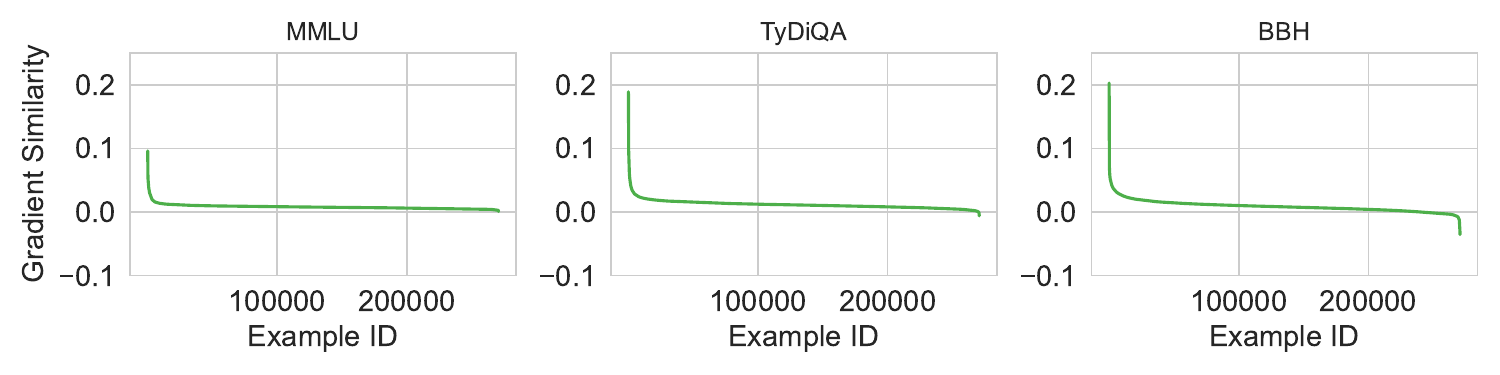}
    \caption{Gradient similarity between training examples and validation examples. The examples are ordered by the gradient similarity score.}
    \label{fig:grads}
\end{figure*}

In~\Cref{fig:grads}, we plot the gradient similarity between each training example and each validation example in descending order. We observe a subset of examples exhibiting relatively higher gradient similarity scores when compared to the validation data. Following this subset, there is a sharp decline in similarity scores, which then continue to decrease in a very smooth manner. Notably, for MMLU, none of the examples exhibit a negative gradient similarity score. In contrast, for TydiQA and BBH, there are instances of examples with negative gradient similarity scores. However, we did not observe any obvious patterns in the examples with negative gradient similarity scores. Please find some examples in \Cref{app:qualitative}. 

\section{Qualitative Analysis}
\label{app:qualitative}
\subsection{Top Selected Examples for \bbh{} and \mmlu{}}
\label{appsub:qual}
We present more selected examples for \mmlu{} and \bbh{} in \Cref{tab:mmlu_example}. For \mmlu{}, the examples are drawn from \cott{}, \cott{}, and \flan{}, respectively. The gradient-based approaches return an example of a problem of a similar difficulty level. BM25 also returns a slightly harder math combinations example, where it has token overlap with the target instance (e.g., multiple digit 5). Representation-based approach return a question answering example that does not require math reasoning ability. For \bbh{}, the examples are drawn from \cott{}, \cott{}, and \dolly{} respectively. The representation and gradient-based approaches capture the essence of the task to judge the truthfulness of a statement, while the BM25 method aligns with superficial ngram features such as NBA Finals.

\subsection{Examples with Negative Gradient Similarity Scores}
\newcommand{\bbhnegtwo}{\textbf{User:} \newline
next question: Where would you find the nearest attic? \newline
Options: \newline
- victorian style house \newline
- basement \newline
- older houses  \newline
- own home \newline
- victorian house \newline
reasoning: \newline
\textbf{Assistant:} \newline
Own home is the house where one lives. One would find the nearest attic in his house. The answer is own home.
}


\newcommand{\bbhneg}{\textbf{User:} \newline
Ques \& Ans: Sam Allardyce has managed the most teams in the Premier League, having taken charge of seven different clubs: Bolton Wanderers, Newcastle United, Blackburn Rovers, West Ham United, Sunderland, Crystal Palace and Everton.
\newline
\newline
Who has managed the most premier league teams?
Sam Allardyce
\newline
\newline
Stream of consciousness: To answer this question, we should know that: Sam Allardyce has managed the most teams in the Premier League, having taken charge of seven different clubs: Bolton Wanderers, Newcastle United, Blackburn Rovers, West Ham United, Sunderland, Crystal Palace and Everton.
\newline
\newline
Ques \& Ans: "Shout" is a song by British band Tears for Fears, written by Roland Orzabal and Ian Stanley and sung by Orzabal (with Curt Smith duetting on the chorus). First released in the UK on 23 November 1984, it was the band's eighth single release (the second taken from their second album Songs from the Big Chair) and sixth UK Top 40 hit, peaking at no. 4 in January 1985. In the US, it reached no. 1 on the Billboard Hot 100 on 3 August 1985 and remained there for three weeks. "Shout" would become one of the most successful songs of 1985, eventually reaching the Top Ten in 25 countries. "Shout" is regarded as one of the most recognisable songs from the mid-eighties and is also recognised as the group's signature song, along with "Everybody Wants to Rule the World".
\newline
\newline
Answer this question: when did shout by tears for fears come out?
23 November 1984
\newline
\newline
Stream of consciousness:
\newline
\textbf{Assistant:} 
The relevant information is: First released in the UK on 23 November 1984, it was the band's eighth single release (the second taken from their second album Songs from the Big Chair) and sixth UK Top 40 hit, peaking at no.}

\newcommand{\tydiqaneg}{\textbf{User:} \newline
Instruction: What would be some useful random items to help you on an island if you were stranded
Output:
\newline
\textbf{Assistant:} \newline
A hockey skate would be great because of the blades and laces
\newline
Glasses could maybe be used to help you start a fire if you used the glasses to reflect the sunlight
\newline
The cord from a blow dryer could be quite useful
}

\newcommand{\tydiqanegthree}{\textbf{User:} \newline
q: Where could you buy an old pie safe with decorative tin?
\newline Options:
\newline - shopping bag
\newline - trash
\newline - antique store
\newline - landfill\newline - cupboard
\newline a: Old items can be found at an antique store. From an antique store, you could buy an old pie safe with decorative tin.
\newline The answer is antique store
\newline QUESTION: What article of apparel or accessory is typically made of felt?
\newline Options:
\newline - shoes
\newline - table
\newline - hat shop
\newline - clothes
\newline - man's hat
\newline ANSWER: Felt is a type of textile material. Man's hat are generally made of felt material. Hat can be article of apparel or accesory.
\newline The answer is man's hat
\newline q: A lot of people play the fiddle in what state which comes first alphabetically?
\newline Options:
\newline - tennessee
\newline - violin case
\newline - orchestra
\newline - alabama
\newline - alaska
\newline a:
\newline \textbf{Assistant:}
\newline Lot of people in alabama state also plays fiddle like other states. Alabama comes first alphabetically than other state as it starts from alphabet A.
\newline The answer is alabama
}

\begin{table*}[ht]
       \caption{Examples with a negative gradient similarity score to the validation examples for \tydiqa{}.}
    \centering
    \begin{tabular}{@{}p{16cm}@{}}
    \toprule
    \multicolumn{1}{c}{\textbf{\tydiqa{} validation example}} \\ \midrule
    \textbf{User:} 
The following are multiple choice questions (with answers) about elementary mathematics.  The population of the city where Michelle was born is 145,826. What is the value of the 5 in the number 145,826? A. 5 thousands B. 5 hundreds C. 5 tens D. 5 ones Answer: \newline \newline 
\textbf{Assistant:} The answer is: A \\ \midrule
\multicolumn{1}{c}{\textbf{Selected examples with lowest $\infadam$ scores}}
 \\ 
\midrule
    \tydiqaneg \\ \midrule
    \tydiqanegthree \\ \bottomrule
    \end{tabular}
    \label{tab:neg_grad_sim_tydiqa}
\end{table*}

\begin{table*}[ht]
    \caption{Examples with a negative gradient similarity score to the validation examples for \bbh{}.}
    \centering
    \begin{tabular}{@{}p{16cm}@{}}
    \toprule
    \multicolumn{1}{c}{\textbf{\bbh{} validation example}} \\ \midrule
    \textbf{User:} \newline
Determine whether an artificially constructed sentence relating to sports is plausible or not.  Q: Is the following sentence plausible? "Bam Adebayo scored a reverse layup in the Western Conference Finals." \newline \newline 
\textbf{Assistant:} A: Let's think step by step. Bam Adebayo is an American basketball player. Scoring a reverse layup in the Western Conference Finals is part of the NBA Finals. So the answer is yes.   \\ \midrule
    \multicolumn{1}{c}{\textbf{Selected examples with lowest $\infadam$ scores}} \\ \midrule
    \bbhneg \\ \midrule
    \bbhnegtwo \\ \bottomrule
    \end{tabular}

    \label{tab:neg_grad_sim_bbh}
\end{table*}

We present a few examples with the lowest gradient similarity scores with the validation data in \Cref{tab:neg_grad_sim_bbh} and \Cref{tab:neg_grad_sim_tydiqa}. Simply looking at the examples fails to reveal any  significant patterns that could explain the observed negative transfer to the target task, which suggests the need for further investigation.

\newcommand{\tydiqaeval}{\textbf{User:} {\bng pRdt/t AdhYaJer tethYr Upr ibhit/t ker inm/nilikht pResh/nr Ut/tr idn.AdhYaJ: baNNGlar nbjagrN {O} bRaHMsmaj-ekin/dRk samaijk-saNNGs/krRitk sNNGs/kar Aaen/daln baNNGlar saNNGs/krRitk {O} Ar/th{oi}nitk jiiben gbhiir pRbhab ibs/tar ker. 1857 saelr mHaibedRaeHr suucna klkatar Aduuer{I} HJeichl. E{I} ibedRaH bYr/th Hel{O} Er pirepRikKet bharetr shasnbhar ekam/painr Hat ethek ibRiTsh rajshik/t sWHes/t gRHN ker. bhart shasenr jnY EkiT bha{I}srJer pd srRiSh/T kra HJ.[11] 1905 sael dhr/miiJ 
\newline pRsh/n: Aibbhk/t sWadhiin baNNGla keb idWkhiN/Dt Hy?} \newline \newline \textbf{Assistant:} {\bng Ut/tr: 1905 sael}}

\newcommand{\tydiqaenglish}{\textbf{User:} Answer the following question based on the content of the given chapter. \newline 
Chapter: The Bengali Renaissance and the Brahmo Samaj - in a nutshell, the social reformers and the reforms they introduced had a profound impact on the social and economic life of Bengal. The beginning of the great rebellion took place in Calcutta in 1857. After the failure of this rebellion, the British Empire accepted the rule of the East India Company's hands as a friendly power. For the governance of India, the position of a Viceroy was created. In 1905, the religious and political motivations led to the division of Bengal [...] \newline  Question: When was Bengal divided? \newline\newline \vspace{2mm}  \textbf{Assistant:} Answer: 1905.}


\newcommand{\tydiqabm}{{\textit{\underline{Masked Word Prediction in Bengali}}}  \vspace{1mm} \newline \textbf{User:} 
Select the most logical word from four options to replace the \textless{}MASK\textgreater{} token in the given Bengali statement. 
 \newline [Q]: Statement: {\bng 1961-62 emousuem eTD eDk/sTaerr entrRtWadhiin {I}NNGerj delr sdsYruuep paiks/tan {O}} \textless{}MASK\textgreater{} {\bng spher OI du{I} eTes/T ANNGshgRHN kern. tar ebailNNG bhNG/giimar kareN eTs/T AibheShek pr/b kKaiNkTa iblimWt HJ.}  Option A: {\bng gLYamargn}~~~Option B: {\bng ilecs/TarshaJaerr}~~~Option C: {\bng bhart} \vspace{2mm} \newline  \textbf{Assistant:} [A]: {\bng bhart} }

\newcommand{\tydiqabmenglish}{{\textit{\underline{Prediction of the Masked Word in Bengali}}}  \vspace{1mm} \newline \textbf{User:} 
Choose the most logical word from four options to replace the \textless{}MASK\textgreater{} token in the given Bengali statement. 
 \newline [Q]: Statement: { In the 1961-62 season, the touring English team's captain participated in the Pakistan \textless{}MASK\textgreater{} sphere for those two tests. His availability for bowling marked the delay of his test debut for some time.}  Option A: { Glamorgan}~~~Option B: { Leicestershire}~~~Option C: { India} \vspace{2mm} \newline  \textbf{Assistant:} [A]: {Glamorgan} }

 \newcommand{\tydiqarep}{{\textit{\underline{Hate Speech Classification in Bengali}}} \vspace{1mm} \newline  \textbf{User:} 
 You are given a hateful post in Bengali that expresses hate or encourages violence towards a person or a group based on the protected characteristics such as race, religion, sex, and sexual orientation. You are expected to classify the post into two classes: personal or non-personal depending on the topic. \newline
 Q: {\bng tuim ik sukhii Het ca{O} na? na ca{I} na ikn/tu ekn? suekh thakel bhuuet iklay Aaim ikl ekhet ca{I} na} \vspace{2mm} \newline \textbf{Assistant: }personal}

 \newcommand{\tydiqarepenglish}{{\textit{\underline{Hate Speech Classification in Bengali}}} \vspace{1mm} \newline  \textbf{User:} 
 You are given a hateful post in Bengali that expresses hate or encourages violence towards a person or a group based on the protected characteristics such as race, religion, sex, and sexual orientation. You are expected to classify the post into two classes: personal or non-personal depending on the topic. \newline
 Q: {Do you not want to be happy alone? Not that I don't want to, but now? When happy, I forget that I don't want to eat alone.} \vspace{2mm} \newline \textbf{Assistant: }personal}

 \newcommand{\tydiqagrad}{{\textit{\underline{Question Answering in English}}} \vspace{1mm}  \newline \textbf{User:} 
Given the question and input, write a reponse to answer the question. Which year was quantum computer demonstrated to be possible? 
\newline
Input: Over the years, experimentalists have constructed small-scale quantum computers using trapped ions and superconductors. In 1998, a two-qubit quantum computer demonstrated the feasibility of the technology, [...]
\newline
Response: \vspace{2mm} \newline 
\textbf{Assistant:} 1998}


\begin{table*}
        \caption{Top selected examples when using one \tydiqa{} example as $\dval$ for data selection with different approaches. The \tydiqa{} example is in the context of Bengali question answering. These examples are drawn from \flan{}, \flan{}, and \dolly{}, respectively. Notably, \ours{} demonstrates the capability to prioritize examples of the same task type (question answering), whereas the other methods tend to be influenced by the linguistic attributes of the validation example, leading to the selection of Bengali examples that do not align with the same task type. Please find the translation of the table in \Cref{tab:tydiqa_example_english}.  }
    \centering
    \small
    \resizebox{\textwidth}{!}{
    \begin{tabular}{p{0.33\columnwidth} p{0.33\columnwidth} p{0.33\columnwidth}}
        \toprule
        \multicolumn{3}{c}{\textbf{A TydiQA Validation Example: Question Answering in Bengali}} \\ \midrule
        \multicolumn{3}{c}{\parbox{\textwidth}{\tydiqaeval}}  \\ \midrule
         \multicolumn{1}{c}{\textbf{Selected by BM25}} & \multicolumn{1}{c}{\textbf{Selected by RDS}} &  \multicolumn{1}{c}{\textbf{Selected by \ours{}}} \\ 
        \midrule
         \tydiqabm & \tydiqarep & \tydiqagrad  \\ \bottomrule
    \end{tabular}}
    \label{tab:tydiqa_example}
\end{table*}

\begin{table*}
        \caption{English translations of \Cref{tab:tydiqa_example}.}
    \centering
    \small
    \resizebox{\textwidth}{!}{
    \begin{tabular}{p{0.33\columnwidth} p{0.33\columnwidth} p{0.33\columnwidth}}
        \toprule
        \multicolumn{3}{c}{\textbf{A TydiQA Validation Example: Question Answering in Bengali (Translated)}} \\ \midrule
        \multicolumn{3}{c}{\parbox{\textwidth}{\tydiqaenglish}}  \\ \midrule
         \multicolumn{1}{c}{\textbf{Selected by BM25 (Translated)}} & \multicolumn{1}{c}{\textbf{Selected by RDS (Translated)}} &  \multicolumn{1}{c}{\textbf{Selected by \ours{}}} \\ 
        \midrule
         \tydiqabmenglish & \tydiqarepenglish & \tydiqagrad  \\ \bottomrule
    \end{tabular}}
    \label{tab:tydiqa_example_english}
\end{table*}

\newcommand{\mmlueval}{\textbf{User:} 
The following are multiple choice questions (with answers) about elementary mathematics.  The population of the city where Michelle was born is 145,826. What is the value of the 5 in the number 145,826? A. 5 thousands B. 5 hundreds C. 5 tens D. 5 ones Answer: \newline \newline 
\textbf{Assistant:} The answer is: A}

\newcommand{\mmlubm}{\textbf{User:} 
Question: A multiple choice test consists of 4 questions, and each question has 5 answer choices. In how many ways can the test be completed if every question is unanswered? Options: (A) 24 (B) 120 (C) 625 (D) 720 (E) 1024 Answer: 5 choices for each of the 4 questions, thus total of $5*5*5*5 = 5^4 = 625$ ways.\newline \newline  \newline  \newline  \textbf{Assistant:} The answer is (C).  }

 \newcommand{\mmlurep}{\textbf{User:} 
 Of course, the most unforgettable classmate I had was also the most vicious bully. I attended a boarding school for badly behaved children and it had its fair share of bullies who had been expelled from mainstream schools. But this particular bully had even the teachers running scared. When he went off on one, it could take up to four teachers or members of care staff to get him under control. [...] At the end of the story, the bully is probably: A. Now a teacher B. Now an article writer C. Deceased from drug abuse D. not enough information  \newline \newline \textbf{Assistant:} C{}}

 \newcommand{\mmlugrad}{\textbf{User:} 
 Q: In this task, you need to provide the correct option for a given problem from the provided options. Problem: the compound ratio of 2 / 3 , 6 / 7 , 4 / 3 and 1 / 8 is given by ? Options: $a)~~2 / 21 , ~~b)~~ 1 / 48 , ~~c)~~ 1 / 98 , ~~d)~~ 1 / 96 , ~~e)~~ 1 / 68$  \newline \newline 
 \textbf{Assistant:}
 a}

 \newcommand{\bbheval}{\textbf{User:} \newline
Determine whether an artificially constructed sentence relating to sports is plausible or not.  Q: Is the following sentence plausible? "Bam Adebayo scored a reverse layup in the Western Conference Finals." \newline \newline 
\textbf{Assistant:} A: Let's think step by step. Bam Adebayo is an American basketball player. Scoring a reverse layup in the Western Conference Finals is part of the NBA Finals. So the answer is yes.  }

\newcommand{\bbhbm}{\textbf{User:} 
Give a stream of consciousness and then the final answer. The 2016 NBA Finals was the championship series of the National Basketball Association (NBA) $2015 -- 16$ season and conclusion of the 2016 playoffs. The Eastern Conference champion Cleveland Cavaliers defeated the Western Conference champion Golden State Warriors $4 -- 3$ in a rematch of the 2015 NBA Finals. [...]
Based on this passage, who did cleveland play in the finals last year?\newline \newline \textbf{Assistant:} To answer this question, we should know that: The Eastern Conference champion Cleveland Cavaliers defeated the Western Conference champion Golden State Warriors $4 -- 3$ in a rematch of the 2015 NBA Finals.
So, the final answer is Golden State Warriors. }

 \newcommand{\bbhrep}{\textbf{User:} 
 Is Y2K relevant to the plot of The Godfather?
 Please give rationale first, then the answer. \newline \newline \newline \newline \newline \newline \newline \newline  \textbf{Assistant:} The story in the Godfather spans from 1945 to 1955. Y2K refers to events related to the formatting and storage of calendar data for dates beginning in the year 2000. The answer is: no.}

 \newcommand{\bbhgrad}{\textbf{User:} 
 Plase answer the following request: Is the following state true or false: Peanuts are vegetables? Answer: \newline \newline  \newline  \newline 
 \textbf{Assistant:}
 This statement is true. Peanuts are part of the legume family and are therefore vegetables.}

\begin{table*}
    \caption{Top examples selected for a \mmlu{} validation instance in the context of elementary mathematics and selected for a \bbh{} target instance in the context of sports understandingusing. We compare examples selected by BM25, RDS and \ours{}.}
    \centering
    \small
    \begin{tabular}{p{8cm}p{8cm}}
        \toprule
        \multicolumn{1}{c}{\textbf{MMLU}} & \multicolumn{1}{c}{\textbf{BBH}}  \\
        \midrule
        \multicolumn{1}{c}{\textbf{A Validation Example}} &  \multicolumn{1}{c}{\textbf{A Validation Example}} \\ 
        \cmidrule(lr){1-1} \cmidrule(lr){2-2}
        \mmlueval & \bbheval \\ \midrule
              \multicolumn{1}{c}{\textbf{Selected by BM25}} &  \multicolumn{1}{c}{\textbf{Selected by BM25}}  \\
              \cmidrule(lr){1-1} \cmidrule(lr){2-2}
        \mmlubm & \bbhbm \\ \midrule
        \multicolumn{1}{c}{\textbf{Selected by RDS}} &  \multicolumn{1}{c}{\textbf{Selected by RDS}}  \\ 
        \cmidrule(lr){1-1} \cmidrule(lr){2-2}
        \mmlurep & \bbhrep \\ \midrule
        \multicolumn{1}{c}{\textbf{Selected by \ours{}}} &  \multicolumn{1}{c}{\textbf{Selected by \ours{}}}  \\  
        \cmidrule(lr){1-1} \cmidrule(lr){2-2}
        \mmlugrad & \bbhgrad \\         \bottomrule
    \end{tabular}
    \label{tab:mmlu_example}
\end{table*}

\subsection{Subset Distribution of Selected Examples}
\label{appsub:subset}

\begin{figure*}[h]
    \centering
    \subfigure[\mmlu]{
        \label{fig:mmlu_dist}
        \includegraphics[width=0.3\columnwidth]{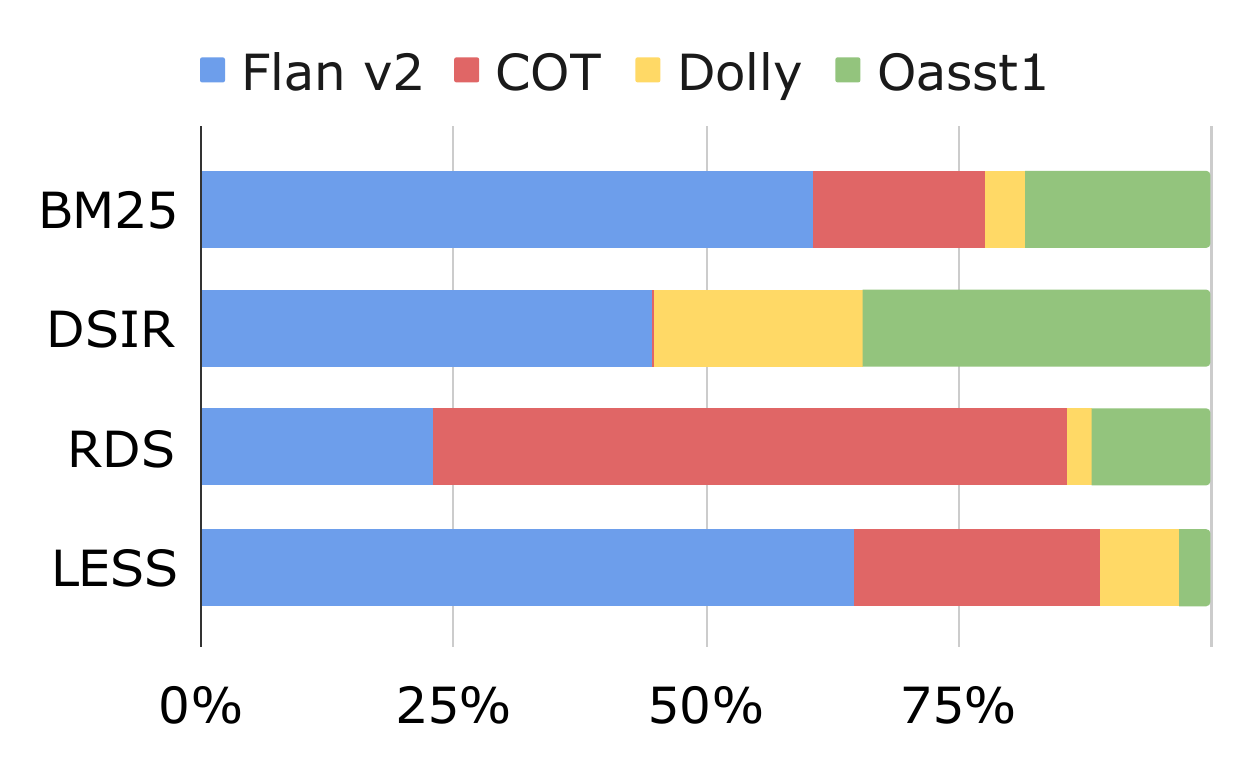}
    }
    \subfigure[\tydiqa]{
        \label{fig:tydiqa_dist}
        \includegraphics[width=0.3\columnwidth]{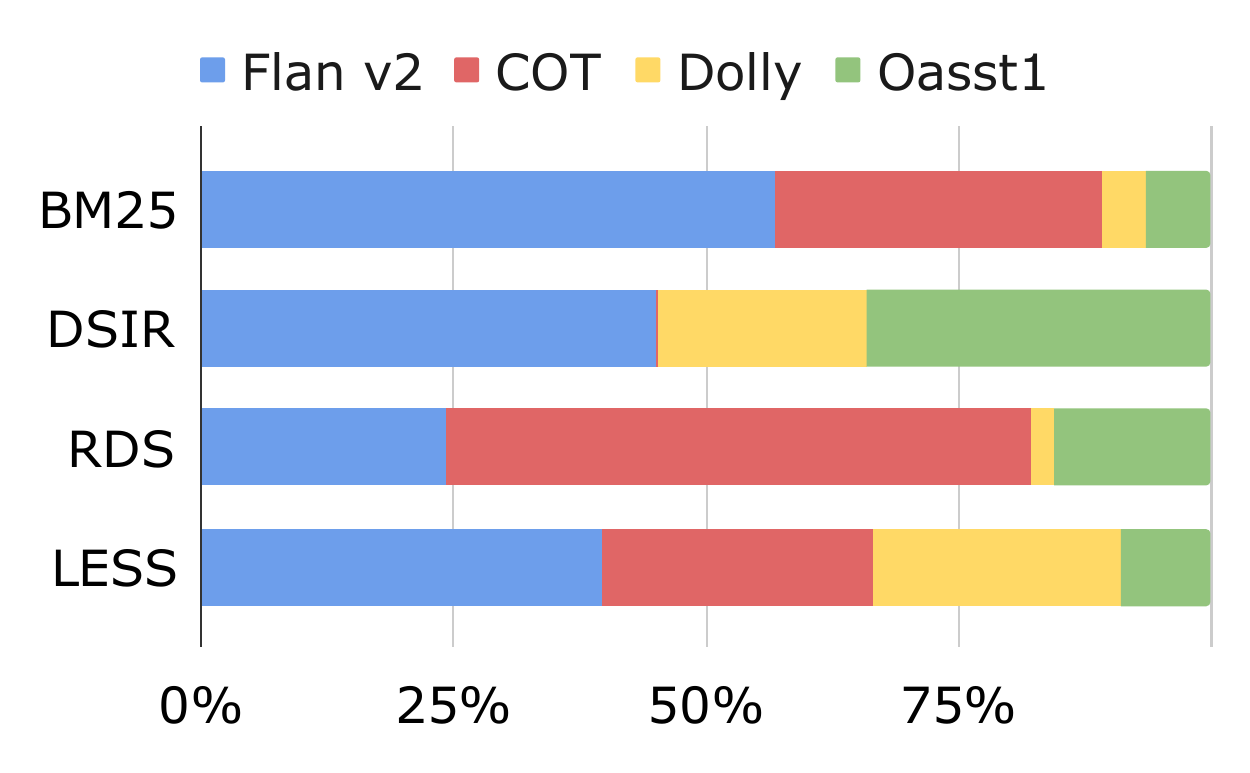}
    }
    \subfigure[\bbh]{
        \label{fig:bbh_dist}
        \includegraphics[width=0.3\columnwidth]{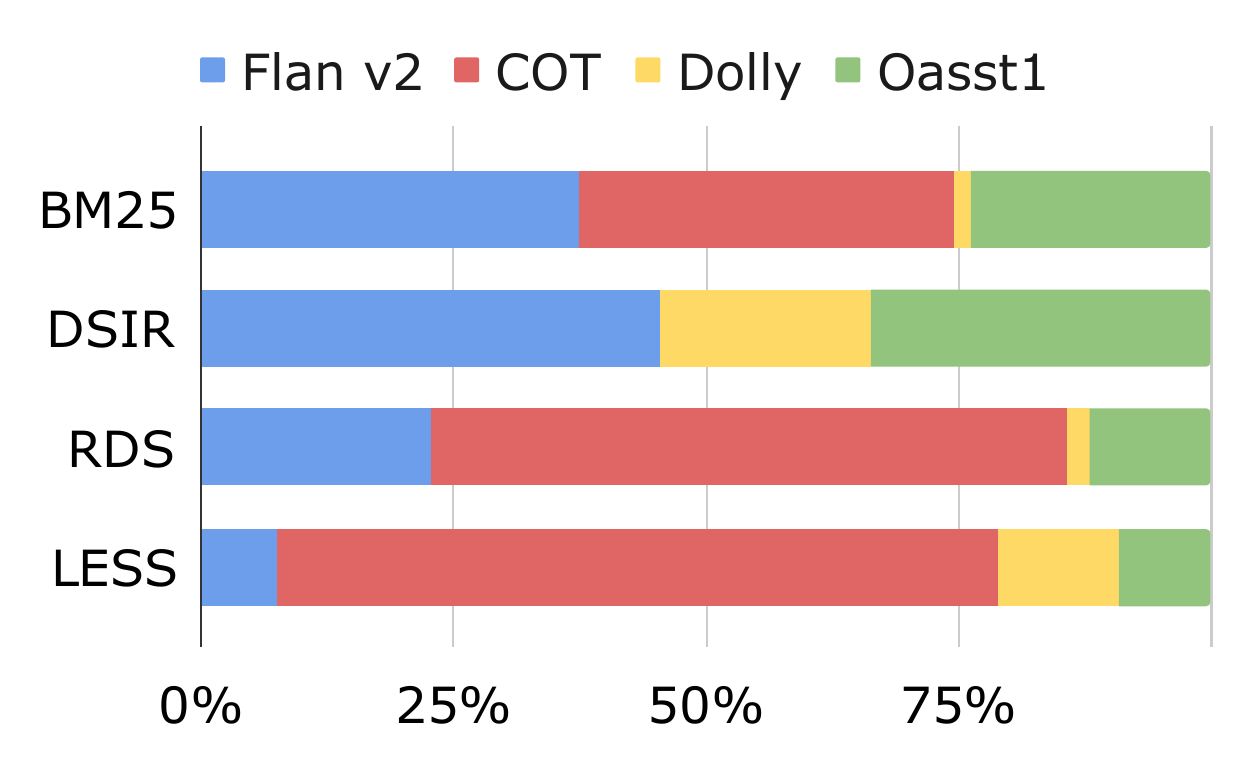}
    }
    \caption{Subset distribution of the top $5\%$ selected examples for different methods.}
    \label{fig:task_dist}   
\end{figure*}

We present the distribution of the selected examples for different methods in \Cref{fig:task_dist}. Different approaches end up selecting very different amounts of data from each training subset. However, we find that there is not a monotonic relationship between the amount of data selected from a subset and the performance of the method. This suggests that each subset could have useful examples for the target task, and the key is to identify the most relevant examples.

In our specific observations, we notice that \ours{} predominantly selects data from the \flan{} dataset for \mmlu{}. This choice appears intuitive since \mmlu{} predominantly comprises multiple-choice questions. In the case of \tydiqa{}, \ours{} exhibits a strong preference for data from the \dolly{} dataset. Upon examination, it becomes evident that many of these examples adhere to an open-book question answering format. Finally, for \bbh{}, \ours{} primarily selects data from the \cott{} dataset. This preference aligns with the fact that \bbh{} tasks involve chain of thought reasoning, a characteristic prominently present in the \cott{} dataset's examples.

\end{document}